\documentclass[11pt]{article}
\usepackage[letterpaper,margin=1in]{geometry}
\usepackage[T1]{fontenc}
\usepackage{lmodern,microtype}
\usepackage{amsmath,amssymb,amsthm,mathtools}
\usepackage{graphicx,booktabs,array}
\usepackage[numbers,sort&compress]{natbib}
\usepackage{tikz}
\usetikzlibrary{arrows.meta,positioning,calc}
\usepackage[font=small,labelfont=bf]{caption}
\usepackage{xcolor}
\definecolor{inkblue}{RGB}{25,61,101}
\definecolor{softblue}{RGB}{235,242,249}
\definecolor{softgray}{RGB}{246,246,246}
\definecolor{warmred}{RGB}{164,62,52}
\usepackage{hyperref}
\hypersetup{colorlinks=true,linkcolor=inkblue,citecolor=inkblue,urlcolor=inkblue,
pdftitle={Omniprediction with the Minimum Number of Linear Scores},
pdfauthor={MohammadHossein Bateni, Zahra Hadizadeh, MohammadTaghi Hajiaghayi, Mahdi JafariRaviz, Shayan Taherijam}}
\theoremstyle{plain}
\newtheorem{theorem}{Theorem}[section]
\newtheorem{binarytheorem}{Theorem}[theorem]

\newtheorem{lemma}[theorem]{Lemma}
\newtheorem{proposition}[theorem]{Proposition}
\newtheorem{corollary}[theorem]{Corollary}
\theoremstyle{definition}
\newtheorem{definition}[theorem]{Definition}
\theoremstyle{remark}

\newcommand{\E}{\mathbb E}
\newcommand{\R}{\mathbb R}
\newcommand{\ip}[2]{\langle #1,#2\rangle}

\DeclareMathOperator{\Lip}{Lip}
\DeclareMathOperator{\prox}{prox}
\DeclareMathOperator{\argmin}{arg\,min}
\title{\bfseries Omniprediction with the Minimum\\Number of Linear Scores}
\author{MohammadHossein Bateni\thanks{Google. Email: \texttt{bateni@google.com}.}\qquad
Zahra Hadizadeh\thanks{University of California, Irvine. Email: \texttt{zhadizad@uci.edu}.}\qquad
MohammadTaghi Hajiaghayi\thanks{University of Maryland. Email: \texttt{hajiagha@umd.edu}.}\\
Mahdi JafariRaviz\thanks{University of Maryland. Email: \texttt{mahdij@umd.edu}.}\qquad
Shayan Taherijam\thanks{University of California, Irvine. Email: \texttt{staherij@uci.edu}.}}
\date{November 2026}
\begin{document}

\maketitle
\begin{abstract}
An $\varepsilon$-omnipredictor uses one forecast for an entire family of downstream losses. After postprocessing for each loss, its expected loss must be at most $\varepsilon$ above that of the best comparator for that loss. A binary single-index model (SIM) learns a norm-bounded weight vector $w$ and a nondecreasing link $\rho$. It reduces the features $x$ to one linear score $w^\top x$, then maps that score to a probability $p(x)=\rho(w^\top x)$. Hu, Tian, and Yang (STOC 2025)~\citep{hty2025} achieved $\varepsilon$-omniprediction for binary matching losses using $O(1/\varepsilon^2)$ SIMs and asked whether one SIM suffices. In the population model, we prove that one SIM suffices for exact omniprediction in their bounded-action setting. The result holds for any compact family of continuous nondecreasing target links.

With unrestricted actions and matching losses generated by $1/\beta$-strongly convex entropies, the SIM's link can also be $\beta$-Lipschitz. For $K$ classes and $d$-dimensional features with finite first moment, $\min\{d,K-1\}$ linear scores are necessary and sufficient in the worst case. A joint link maps these scores to class probabilities, and its Lipschitz bound $\beta$ is also sharp under the comparator's coefficient norm bound.

For bounded features and smooth entropies, we achieve $\varepsilon$-omniprediction for two or three classes using $\widetilde O(1/\varepsilon^2)$ samples at fixed confidence and regularity. This bound is independent of $d$ and tight up to logarithmic factors. For any fixed number of classes $K \ge 2$, the tight worst-case sample complexity for $\varepsilon$-omniprediction under the Dirichlet condition on the features is $\Theta_K(1/\varepsilon^2)$.
\end{abstract}

\section{Introduction}
\label{sec:intro}
An omnipredictor is learned before its downstream loss is specified.
After a loss is chosen, a Bayes decision rule converts the common
probability forecast into an action. The resulting risk must be competitive
with the best member of a prescribed comparator class for that loss.
The same forecast must support all these comparisons
\citep{gopalan2022}. This requirement asks more of the representation than
ordinary loss minimization: different losses may have different optimal
linear predictors, yet the forecast must retain enough information to
serve all of them.

For binary labels, a particularly natural representation is a monotone
single-index model,
\[
                  p(x)=\rho(w^\top x).
\]
The vector $w$ assigns one score to each feature vector, and the
nondecreasing link $\rho$ turns that score into a probability.
For $K$ classes, the corresponding representation is a joint map
of several linear scores into the probability simplex. The question is
whether omniprediction requires more scores than a predictor designed
for one loss.

Hu, Tian, and Yang~\citep{hty2025} made substantial progress on this
question. Their efficient binary omnipredictors use approximately
$\varepsilon^{-2}$ single-index components, and they ask whether
one monotone single-index model can suffice. We study this representation
question for global matching losses, whose downstream action can be any
real score vector, and for binary matching losses whose downstream action
must lie in the same bounded interval as the linear comparators.

\subsection{Main theorems}
\label{sec:main-results}
We start with binary existence, give the multiclass theorem and its
matching score lower bound, and then state the sample upper and lower
bounds. Their significance is discussed in Section~\ref{sec:significance}.

\paragraph{Binary matching losses on the real line.}

Fix an integer $d\ge0$. For $\beta>0$, let $\varphi:[0,1]\to\R$ be continuous
and differentiable on $(0,1)$, with
$q\mapsto\varphi(q)-q^2/(2\beta)$ convex on $[0,1]$.
For actions $a\in\R$ and labels $y\in\{0,1\}$, define
\[
 g_\varphi(a)=\max_{q\in[0,1]}\{qa-\varphi(q)\},
 \qquad \ell_\varphi(a,y)=g_\varphi(a)-ya.
\]
For $0<p<1$, put
$L_\varphi(p,y)=-\varphi(p)+\varphi'(p)(p-y)$.
At an endpoint, set $L_\varphi(p,y)=-\varphi(p)$ if $p=y$ and
$L_\varphi(p,y)=+\infty$ otherwise.

\stepcounter{theorem}
\begin{binarytheorem}[Monotone single-index existence]
\label{thm:binary}
Let $(X,Y)$ be a measurable random pair in $\R^d\times\{0,1\}$ with
$\E\|X\|<\infty$, and let $R,\beta>0$. There are $w\in\R^d$ with
$\|w\|_2\le R$ and a nondecreasing $\beta$-Lipschitz map
$\rho:\R\to[0,1]$ such that, simultaneously for every $\varphi$ above,
\[
 \E L_\varphi(\rho(w^\top X),Y)
 \le\inf_{\|v\|_2\le R}\E\ell_\varphi(v^\top X,Y).
\]
The forecast loss and every comparator loss are integrable, and the
comparator infimum is finite. Almost surely,
$\rho(w^\top X)=0$ implies $Y=0$, and $\rho(w^\top X)=1$ implies $Y=1$.
The same conclusion holds with any nonempty compact convex set of
weights in place of the Euclidean ball, in both the forecast and
comparator constraints.
\end{binarytheorem}

\paragraph{Bounded binary actions.}
Let $L,R>0$, $A=LR$, and $I=[-A,A]$. For a continuous nondecreasing
target link $\sigma:I\to[0,1]$, put
\[
 G_\sigma(t)=\int_0^t\sigma(s)\,ds,\qquad
 \ell_\sigma(t,y)=G_\sigma(t)-yt,\qquad
 k_\sigma(q)\in\argmin_{t\in I}\{G_\sigma(t)-qt\}.
\]
Fix a choice of Bayes action at ties. Every such choice is nondecreasing
in $q$ and hence Borel measurable.

\begin{binarytheorem}[One score with bounded actions]\label{thm:bounded-binary}
Suppose $X\in\R^d$, $\|X\|_2\le L$ almost surely, and $Y\in[0,1]$.
Let $\mathcal S$ be a nonempty family of continuous nondecreasing links
$I\to[0,1]$, compact in the uniform norm. There exist a vector $w\in\R^d$ with $\|w\|_2\le R$ and
a nondecreasing Borel map $\rho:\R\to[0,1]$ such that $p(x)=\rho(w^\top x)$
satisfies, simultaneously for every $\sigma\in\mathcal S$,
\[
 \E\ell_\sigma(k_\sigma(p(X)),Y)
 \le \inf_{\|v\|_2\le R}\E\ell_\sigma(v^\top X,Y).
\]
\end{binarytheorem}
The target family may consist of all nondecreasing $\beta$-Lipschitz
links on $I$ for any fixed finite $\beta$. The learned link in
Theorem~\ref{thm:bounded-binary} has no prescribed slope bound.

\paragraph{Multiclass matching losses.}
Fix integers $K\ge2$ and $d\ge0$, and let $R,\beta>0$.
Let $(X,Y)$ be a measurable random pair with an arbitrary joint law on
$\R^d\times[K]$. Let $\Delta_K$ be the
probability simplex, put $u=\mathbf1/K$ and $H=\mathbf1^\perp$, and
use Euclidean norms on vectors and the Frobenius norm on coefficient
matrices. For each $y\in[K]$, let $e_y$ denote the $y$th standard basis vector of $\mathbb R^K$. Let $\mathcal E_\beta$ be the family of finite continuous
$1/\beta$-strongly convex entropies on $\Delta_K$ that are
differentiable on the relative interior of every face. For
$\Phi\in\mathcal E_\beta$, define
\[
 G_\Phi(a)=\max_{q\in\Delta_K}\{\langle q-u,a\rangle-\Phi(q)\},
 \qquad
 \ell_\Phi(a,y)=G_\Phi(a)-\langle e_y-u,a\rangle,
 \quad a\in H.
\]
The comparator actions are $VX$, where $V:\R^d\to H$ and
$\|V\|_F\le R$. Downstream actions are unrestricted real score
vectors. For $J=\operatorname{supp}(p)$, let $\nabla_J\Phi(p)$ be
the gradient tangent to the face $\Delta_J$, and set
\[
 L_\Phi(p,y)=
 \begin{cases}
 -\Phi(p)+\langle\nabla_J\Phi(p),p-e_y\rangle,&y\in J,\\
 +\infty,&y\notin J.
 \end{cases}
\]
On the interior this is the loss of the canonical Bayes action
$\nabla_H\Phi(p)$. Section~\ref{sec:model} gives examples and
explains the boundary convention.

\begin{theorem}[Multiclass existence]
\label{thm:main}\label{core:global}
Assume $\E\|X\|<\infty$. For every $R,\beta>0$, there exist a linear map
$W:\R^d\to H$ with $\|W\|_F\le R$ and a $\beta$-Lipschitz map
$\rho:H\to\Delta_K$ such that $p=\rho\circ W$ uses at most
$\min\{d,K-1\}$ independent linear scores and, simultaneously for every
$\Phi\in\mathcal E_\beta$,
\[
 \E L_\Phi(p(X),Y)
 \le \inf_{\substack{V:\R^d\to H\ \mathrm{linear}\\\|V\|_F\le R}}
       \E\ell_\Phi(VX,Y).
\]
For each such $\Phi$, the forecast loss and every comparator loss are
integrable, and the displayed infimum is finite. Moreover, for every
$j\in[K]$, $\mathbb P(p_j(X)=0,\,Y=j)=0$.
More generally, any nonempty compact convex set $\mathcal W$ of linear
score maps may replace the norm ball in both the forecast and comparator
constraints. This includes closed Frobenius- and operator-norm balls
of radius $R\ge0$, and their intersections with additional closed
convex coefficient constraints, provided the intersection is nonempty.
\end{theorem}

The next theorem gives the exact cost of restricting the number of
scores or the Lipschitz bound. Write $(a)_+=\max\{a,0\}$. Regret
means excess expected loss over the best radius-$R$ linear action.

\begin{theorem}[Joint score--slope tradeoff]
\label{thm:rank}\label{prop:slope}
Fix integers $K\ge2$ and $d\ge1$, let $R,\beta,L>0$, and put
$d_\star=\min\{d,K-1\}$. Suppose $\beta RL/\sqrt{d_\star}<1/K$.
There is one joint law of $(X,Y)\in\R^d\times[K]$, with $X$ uniform
on the open radius-$L$ ball centered at the origin in a $d_\star$-dimensional
linear subspace, and one quadratic entropy in $\mathcal E_\beta$ with
the following property. Simultaneously for every integer $1\le r\le d_\star$
and every $B\ge0$, consider forecasts $p=\rho\circ W$ with linear
$W:\R^d\to H$ satisfying
\[
 \operatorname{rank}(W)\le r,\qquad
 \|W\|_F\le R,\qquad
 \rho:H\to\Delta_K,\qquad\operatorname{Lip}(\rho)\le B.
\]
Among these forecasts the minimum regret is
\begin{equation}
 \frac{L^2}{2\beta(d_\star+2)}
 \left[(d_\star-r)c^2+r\left(c-\frac{BR}{\sqrt r}\right)_+^2\right],
 \label{eq:joint-main}
\end{equation}
where $c=\beta R/\sqrt{d_\star}$. For $r=0$ and every $B\ge0$, the value is
$L^2d_\star c^2/(2\beta(d_\star+2))$.
Here regret is $\E L_\Phi(p(X),Y)$ minus the finite benchmark
$\inf_{\|V\|_F\le R}\E\ell_\Phi(VX,Y)$ over linear $V:\R^d\to H$.
Every displayed value is attained. Uniform measure on the feature ball
means normalized $d_\star$-dimensional Lebesgue measure in its support subspace.
\end{theorem}
For $r<d_\star$ the omitted directions give strictly positive regret, even if
the link is only required to be measurable (Appendix~\ref{sec:core-appendix}).
Together with Theorem~\ref{thm:main}, this gives the exact worst-case
score count $d_\star=\min\{d,K-1\}$.

\paragraph{General sample upper and lower bounds.}
For the learning results below, assume $K\ge2$, $L,R,\beta,\lambda,M>0$, and $\|X\|\le L$ almost surely, and put $A=LR$. Let $\mathcal E(\beta,\lambda,M)$ contain the entropies in $\mathcal E_\beta$ admitting a convex $C^1$ extension to a convex relatively open neighborhood of $\Delta_K$ in its affine hull. Their relative gradients lie in $H$ and satisfy, for all $q,q'\in\Delta_K$ in Euclidean norm,
\begin{align}
 \|\nabla_H\Phi(q)-\nabla_H\Phi(q')\|
       &\le\lambda\|q-q'\|,\notag\\
 \sup_{q\in\Delta_K}\|\nabla_H\Phi(q)\|&\le M.
 \label{eq:smooth-family}
\end{align}
We now use the finite canonical Bayes action $k_\Phi(q)=\nabla_H\Phi(q)$ and write $\mathcal L_\Phi(q,y)=\ell_\Phi(k_\Phi(q),y)$, including at the boundary. This is a valid Bayes minimizer by Fenchel equality. It need not equal the extended boundary convention above on an excluded label. The finite action is necessary for the uniform statistical guarantee below. In particular, untrimmed Shannon entropy is not covered by this statistical theorem.

\begin{theorem}[Ambient-dimension-free learning]
\label{thm:pac}\label{stat:pac}
Fix an integer $d\ge0$, and put $m=K-1$ and $D=M+A+\beta\lambda A$. For every integer $n\ge1$ satisfying the sample bound below, there is a deterministic finite-search learner, chosen independently of the data law, returning $p(x)=\rho(Wx)$ with linear $W:\mathbb R^d\to H$, $\|W\|_F\le R$, and $\rho:H\to\Delta_K$ of Lipschitz constant at most $\beta$. Its prediction rule is jointly Borel measurable in the sample and query. For every Borel probability law of $(X,Y)\in\mathbb R^d\times[K]$ with $\|X\|\le L$ almost surely, with probability at least $1-\delta$ over the i.i.d. sample,
\begin{align}
 &\E\ell_\Phi(k_\Phi(p(X)),Y)\notag\\
 &\hspace{5mm}\le\inf_{\|V\|_F\le R}\E\ell_\Phi(VX,Y)+\varepsilon
 \label{eq:pac-main}
\end{align}
simultaneously for $\Phi\in\mathcal E(\beta,\lambda,M)$. For $0<\varepsilon\le D$ and $0<\delta<1/2$, sufficient sample sizes are
\begin{equation}
 n\ge
 \begin{cases}
 C_K(D/\varepsilon)^2\log^2\!\dfrac{C_KD}{\varepsilon\delta},&m\le2,\\[1mm]
 C_K\big[(D/\varepsilon)^m+(D/\varepsilon)^2
       \log\dfrac2\delta\big],&m\ge3.
 \end{cases}
 \label{eq:sample-main}
\end{equation}
Here $C_K\ge1$ depends only on $K$. No general polynomial running time is asserted.
\end{theorem}

\begin{proposition}[Quadratic sample lower bound under spherical symmetry]
\label{stat:lower}
Fix $K\ge2$ and $L=R=\beta=\lambda=M=1$. Even for the single quadratic
entropy $\Phi(q)=\|q-u\|^2/2$, any learner that guarantees
\eqref{eq:pac-main} with success probability greater than $5/7$ needs
$\Omega_K(\varepsilon^{-2})$ samples in the worst case. The lower bound
allows randomized learners and arbitrary measurable output forecasts.
In the construction, $X$ is uniform on a Euclidean unit sphere and
$d=\Theta_K(1/\varepsilon)$.
\end{proposition}

\paragraph{Matching sample bounds on finite dictionaries.}
For a class of finite feature dictionaries, the quadratic sample exponent
holds for every fixed $K$. Call a finite dictionary $x_1,\ldots,x_N\in\R^d$, with
$\|x_i\|\le L$, \emph{Dirichlet} if its points are affinely independent
and the pseudoinverse of its centered Gram matrix has nonpositive
off-diagonal entries. The explicit matrix condition and examples appear
in Section~\ref{sec:dirichlet-learning}. No restriction is imposed on
the feature probabilities or conditional labels. Write $\Pi_C$ for
Euclidean projection onto a closed convex set $C$.

\begin{theorem}[Quadratic sample bound on Dirichlet dictionaries]
\label{thm:dirichlet-learning}
Assume the feature law is supported on a Dirichlet dictionary of $N\ge1$ points in $\mathbb R^d$, where $d\ge0$, and put
$D=M+LR+\beta\lambda LR$. For $0<\varepsilon\le D$ and
$0<\delta<1/2$, there is a learner using
\begin{equation}
 n\ge c\frac{D^2}{\varepsilon^2}\log\frac2\delta
 \label{eq:dirichlet-samples}
\end{equation}
i.i.d. examples and returning
\[
 \widehat p(x)=u+\Pi_{\Delta_K-u}
                  (\widehat a+\beta\widehat W x),\qquad
 \|\widehat W\|_F\le R,\quad \widehat a\in H,
\]
such that, with probability at least $1-\delta$, simultaneously for
every $\Phi\in\mathcal E(\beta,\lambda,M)$,
\[
 \E\ell_\Phi(k_\Phi(\widehat p(X)),Y)
 \le\inf_{\|V\|_F\le R}\E\ell_\Phi(VX,Y)+\varepsilon.
\]
Here $c$ is a numerical constant independent of $K,d,N$. The forecast
uses at most $K-1$ scores and a $\beta$-Lipschitz joint link whose
centered version is the gradient of a differentiable convex function.
The learner is deterministic and uses only the sample and the numerical parameters.
\end{theorem}

\begin{proposition}[Optimal accuracy exponent on Dirichlet dictionaries]
\label{prop:dirichlet-lower}
For every fixed $K\ge2$, a learner satisfying the conclusion of
Theorem~\ref{thm:dirichlet-learning} with success probability greater
than $5/7$ requires $\Omega_K(\varepsilon^{-2})$ observations in the
worst case. This holds for the single quadratic entropy, fixed
$L=R=\beta=\lambda=M=1$, and unrestricted output probability
forecasts. The dictionary size and ambient dimension are
$\Theta_K(1/\varepsilon)$.
\end{proposition}

\subsection{Significance and comparison}
\label{sec:significance}

\paragraph{An exact representation theorem.}
Theorems~\ref{thm:binary} and~\ref{thm:main} show that the number
of scores need not grow with inverse accuracy. They give an exact
population guarantee with the original norm and Lipschitz budgets.
The coefficient constraint can even be an arbitrary nonempty compact
convex set. A single deterministic forecast supports every downstream
loss, without knowing which loss will be selected.

The theorem bounds the linear information that the forecast must retain.
An arbitrary $K$-class conditional probability vector
need not be determined by $K-1$ linear statistics of the features.
The theorem shows that these statistics nevertheless suffice to match
every norm-bounded linear comparator, simultaneously over the loss
family. The conclusion does not require the labels to follow a
single-index or multiclass linear model.

\paragraph{Matching lower bounds.}
The score count is best possible. One realizable quadratic instance gives
the \emph{exact} minimum regret when the number of scores or the
Lipschitz bound is reduced (Theorem~\ref{thm:rank}). The regret separates
into the variance of the discarded directions and the remaining error
caused by a slope bound that is too small. The score lower bound continues
to hold for arbitrary measurable links, with no continuity assumption.
Thus the upper and lower bounds determine the worst-case score requirement
for this model in every feature dimension.

The multiclass link must be allowed to use its scores jointly. On a
realizable softmax instance, constraining even one class probability to
be a function of one linear score forces a positive log-loss gap
(Theorem~\ref{separate:impossibility}). This concerns separate probability
links; normalized logits, including ordinary softmax, already form a
joint probability map.

\paragraph{Learning the same representation.}
For bounded features and smooth entropies, a finite learner retains the
same score count, norm, and Lipschitz bound, with sample complexity
independent of $d$ (Theorem~\ref{thm:pac}). At fixed regularity and
confidence, the bound is $\widetilde O_K(\varepsilon^{-2})$ for
$K=2,3$ and $O_K(\varepsilon^{-(K-1)})$ for larger $K$.
Proposition~\ref{stat:lower} shows that the quadratic exponent is necessary
already for one loss when the feature dimension may grow. The optimal
exponent for general features with $K\ge4$ remains open.

On a finite feature dictionary whose
squared minimum affine slope norm is a graph Dirichlet energy,
ordinary convex least squares followed by simplex
projection gives a joint convex-gradient link and
$O(\varepsilon^{-2}\log(1/\delta))$ samples for every $K$ at fixed
regularity (Theorem~\ref{thm:dirichlet-learning}). A one-hot dictionary
lower bound shows that this accuracy exponent is optimal for each
fixed $K$, even without restricting the learner's output.

\begin{table}[t]
\centering
\small
\setlength{\tabcolsep}{5pt}
\renewcommand{\arraystretch}{1.18}
\begin{tabular}{@{}>{\raggedright\arraybackslash}p{0.25\textwidth}>{\raggedright\arraybackslash}p{0.27\textwidth}>{\raggedright\arraybackslash}p{0.40\textwidth}@{}}
\toprule
Setting & Representation & Main guarantee\\
\midrule
Global actions, finite first moment & One binary score with a monotone link or $\min\{d,K-1\}$ joint scores
& Exact population comparison with prescribed norm and slope\\
Bounded binary actions and features & One score with a monotone link
& Exact comparison for a compact family of continuous monotone target links\\
Bounded features, smooth entropies & Same norm, score count, and slope as for global actions
& Dimension-free sample upper bound. The quadratic lower bound matches up to logs for $K=2,3$\\
Dirichlet dictionary & At most $K-1$ scores with a joint projection link
& Matching quadratic sample exponent for every fixed $K$\\
\bottomrule
\end{tabular}
\caption{Representation and sample guarantees. The lower bounds on sample
size are worst-case bounds at fixed high confidence and allow the
feature dimension to grow. All rows allow arbitrary conditional labels.}
\label{tab:overview}
\end{table}

\paragraph{The improvement and its scope.}
The central improvement is an exact, sharp bound on the linear
representation. It is not a general improvement in the sample complexity
of unrestricted omnipredictors. In the global-action model,
exact boundary comparisons use specified extended losses, with finite
actions attaining every positive accuracy for each target.
The general existence theorem supplies a joint Lipschitz multiclass
link, but does not require it to be a convex gradient. General efficient
learning under an arbitrary feature marginal is not established.
These distinctions delimit the representation question settled by the
upper and lower bounds.

\subsection{Related work}
\paragraph{Omniprediction and statistical complexity.}
Omniprediction was introduced by Gopalan et al.~\citep{gopalan2022}.
Outcome indistinguishability and calibrated multiaccuracy provide
sufficient conditions for matching-loss omniprediction
\citep{gopalan2023}. Our proof uses a weaker, one-sided certificate
derived from least-squares optimality. Near-optimal algorithms for broad
loss classes, deterministic omniprediction for proper losses, and optimal
deterministic multicalibration have since been developed
\citep{okoroafor2025,gibbs2026,noarov2026}. Those results do not impose
the linear-score architecture studied here. Properness of a loss and
membership of a forecast in a prescribed hypothesis class are separate
notions. Omniprediction for regression and panprediction address further
forms of reusable prediction~\citep{gopalan2024,balakrishnan2026}.

\paragraph{Single-index and multiclass models.}
The direct comparison is with~\citet{hty2025}: their monotone-Lipschitz
binary learner uses approximately $\varepsilon^{-4}$ samples,
improving to $\varepsilon^{-2}$ for bi-Lipschitz links, and outputs
approximately $\varepsilon^{-2}$ heads. Their Bayes actions are bounded.
Theorem~\ref{thm:bounded-binary} gives an exact bounded-action existence
result for each compact continuous target family, including the full
family with any common finite target slope. Its learned link is
nondecreasing and Borel. Theorem~\ref{thm:binary} also preserves the
prescribed slope of the learned link for global actions.
Earlier work studies multidimensional predictions for downstream decisions
\citep{noarov2025} and omniprediction for nonlinear losses~\citep{lu2025}.
For multiclass omniprediction, \citet{hty2026} obtain approximately
$\varepsilon^{-(K+1)}$ samples in a different, broader loss model.
They also exhibit a failure of a natural loss-independent multiclass
isotonic construction under cyclic-monotonicity constraints.
Our general link is permitted to be any joint Lipschitz simplex-valued
map. The convex-gradient conclusion in the Dirichlet theorem relies
on its additional feature geometry.

\paragraph{Regression, convex losses, and the tools we use.}
Isotron and its Lipschitz version are the algorithmic predecessors of
our scalar fits~\citep{kalai2009,kakade2011}; agnostic single-index
learning has also been studied through omniprediction and alignment
arguments~\citep{gollakota2023,zarifis2024}. Ordinary isotonic regression
has simultaneous optimality properties for proper scoring rules
\citep{brummer2013,jordan2022}. Our global-action binary theorem preserves a prescribed slope
bound and compares against norm-bounded linear actions.
For bounded-norm linear prediction under squared loss,
\citet{shamir2015} gives tight agnostic sample complexity bounds.
Our lower bounds show that quadratic dependence on the inverse accuracy
persists for unrestricted probability forecasts under the feature
geometries considered here.
Our conjugate matching losses are Fenchel--Young losses up to
label-only constants~\citep{blondel2020}. The explicit extension of
fitted vector-valued probabilities uses Kirszbraun's theorem and its
finite-data formula~\citep{kirszbraun1934,azagra2021}.
The simplex projection used by the Dirichlet learner is the sparsemax
map~\citep{martins2016}.

\paragraph{Organization.}
Section~\ref{sec:model} gives further model details and examples.
Section~\ref{sec:techniques} explains the population upper and lower
bounds. Sections~\ref{sec:statistics} and~\ref{sec:dirichlet-learning}
develop the sample bounds for general bounded features and Dirichlet
dictionaries. Appendices~\ref{sec:binary-appendix}
and~\ref{sec:core-appendix} give the binary and multiclass existence
proofs and the sharp score count. Appendix~\ref{separate:section}
explains why multiclass probabilities can require a joint link.
Appendices~\ref{sec:statistics-appendix} and~\ref{app:dirichlet}
prove the statistical results.

\section{Model, Examples, and Consequences}
\label{sec:model}
\subsection{The scalar binary normalization}
\label{sec:binary}
The global binary theorem uses the uncentered scalar convention, which we recall here. Let $Y\in\{0,1\}$ and let $\varphi:[0,1]\to\R$ be continuous, differentiable on $(0,1)$, and $1/\beta$-strongly convex. Define
\[
 g_\varphi(a)=\max_{q\in[0,1]}\{qa-\varphi(q)\},
 \qquad \ell_\varphi(a,y)=g_\varphi(a)-ya.
\]
For $p\in(0,1)$, set $L_\varphi(p,y)=-\varphi(p)+\varphi'(p)(p-y)$; at $p=0,1$, use the endpoint loss in Theorem~\ref{thm:binary}.

The weight ball may likewise be replaced by any nonempty compact convex weight set. In particular, prescribed coefficient signs can be retained.
The proof uses monotone Lipschitz fits. Scalar clipping, translations followed by clipping, and proximal maps of convex functions are nondecreasing and nonexpansive. The own-score perturbation is also nondecreasing. Thus the entire cancellation argument stays inside the monotone SIM class; Appendix~\ref{sec:binary-appendix} gives the details. This scalar convention has its own norm and slope normalization.

Under the bounded-action convention of Theorem~\ref{thm:bounded-binary},
the decision maker minimizes the reported loss over $[-LR,LR]$.
Both binary forecasts depend on only one linear score. With bounded
actions, the compactness assumption concerns the target links, and the
learned monotone link may have jumps.

\subsection{The multiclass model}
Let $K\ge2$, let $Y\in[K]$, let $u=\mathbf 1/K$, and write
\[
 H=\mathbf1^\perp\subset\R^K,\qquad Z=e_Y-u,
 \qquad m=K-1.
\]
Here $Z$ is the centered label vector, and $\Delta_K$ is the probability simplex. A comparator is a linear map $V:\R^d\to H$ with $\|V\|_F\le R$; the population existence theorem permits any nonempty compact convex coefficient set. A forecast has the form
\begin{equation}
 p(x)=\rho(Wx),\qquad \rho:H\to\Delta_K.
 \label{eq:architecture}
\end{equation}
An orthonormal identification of $H$ with $\R^m$ expresses this architecture using $m$ real scores. The restriction to linear scores is substantive: the dimension of a probability vector alone does not imply that it factors through $m$ linear statistics of the features. The link is joint: each class probability may depend
on all scores. Neither averaging the scores nor applying separate scalar
probability links is part of the model.

\begin{definition}[Entropy family]
\label{def:entropy}
For $\beta>0$, let $\mathcal E_\beta$ consist of finite continuous
functions $\Phi:\Delta_K\to\R$ that are differentiable on the relative
interior of every nonempty face and satisfy
\begin{equation}
 q\longmapsto \Phi(q)-\frac{1}{2\beta}\|q-u\|^2
 \quad\text{is convex on }\Delta_K.
 \label{eq:strong}
\end{equation}
All norms here are Euclidean. For a nonempty $J\subseteq[K]$, face
differentiability means differentiability of the restriction to
$\Delta_J$ along
$H_J=\{v\in H:v_j=0\text{ for }j\notin J\}$.
Its relative gradient is $\nabla_J\Phi(p)\in H_J$; on a singleton face
it is zero. We write $\nabla_H\Phi=\nabla_{[K]}\Phi$ on
$\operatorname{relint}\Delta_K$.
\end{definition}
Define the global matching potential and loss by
\begin{align}
 G_\Phi(a)&=\max_{q\in\Delta_K}
       \{\ip{q-u}{a}-\Phi(q)\},\quad a\in H,
       \label{eq:potential}\\
 \ell_\Phi(a,y)&=G_\Phi(a)-\ip{e_y-u}{a}.
 \label{eq:matching}
\end{align}
The maximizing probability is unique, and $u+\nabla G_\Phi$ is a $\beta$-Lipschitz simplex-valued link. Moreover, $G_\Phi$ is $1$-Lipschitz. Adding the label-only term $\Phi(e_y)$ gives the usual Fenchel--Young loss~\citep{blondel2020}. Comparator actions are $VX$; the downstream Bayes action is not constrained to a ball of radius $R\|X\|$.

\paragraph{Boundary convention.}
For $p$ with support $J$, let $\nabla_J\Phi(p)$ be the gradient tangent to the face $\Delta_J$. We use the extended canonical loss
\begin{equation}
 L_\Phi(p,y)=
 \begin{cases}
 -\Phi(p)+\ip{\nabla_J\Phi(p)}{p-e_y},&y\in J,\\
 +\infty,&y\notin J.
 \end{cases}
 \label{eq:extended}
\end{equation}
On the interior this equals $\ell_\Phi(\nabla_H\Phi(p),y)$.
On a face it is obtained as a limit of matching losses by sending
excluded-class scores to $-\infty$ relative to the retained scores.
The loss is bounded below by $-\Phi(e_y)$.
It is Borel measurable: on each relative face interior the finite
formula is continuous, since the relative gradient of a differentiable
convex function is continuous, and the excluded-label branch is
$+\infty$ on a Borel set. Appendix~\ref{sec:core-appendix} justifies these statements. Theorem~\ref{thm:main} guarantees both that an excluded class never occurs and that all loss expectations in the comparison are finite. Finite downstream actions can approximate these comparisons without retraining: replace $p$ by $(1-t)p+tu$ only inside the action rule and use $\nabla_H\Phi((1-t)p+tu)$. Corollary~\ref{revision:finite-actions} bounds the resulting error and shows that it vanishes as $t\downarrow0$ for each target entropy. The smoothing parameter may depend on the target; no common finite-action bound is asserted over the entire entropy family.

\begin{definition}[Exact matching-loss omniprediction]
Let $(X,Y)$ be a measurable random pair in $\R^d\times[K]$, and let
$R\ge0$. A Borel-measurable forecast $p:\R^d\to\Delta_K$ exactly
omnipredicts $\mathcal E_\beta$ against the radius-$R$ comparator ball
if, simultaneously for every $\Phi\in\mathcal E_\beta$,
\begin{equation}
 \E L_\Phi(p(X),Y)
 \le \inf_{\substack{V:\R^d\to H\ \mathrm{linear}\\\|V\|_F\le R}}
       \E\ell_\Phi(VX,Y).
 \label{eq:omni}
\end{equation}
Expectations are ordinary extended Lebesgue expectations. Both losses
are bounded below by $-\max_{y\in[K]}\Phi(e_y)$, so each expectation
is well defined in $\R\cup\{+\infty\}$.
An $\varepsilon$-guarantee, for $\varepsilon\ge0$, adds $\varepsilon$
to the right-hand side.
\end{definition}

\paragraph{Examples.}
The quadratic entropy $\Phi(q)=\|q-u\|^2/(2\beta)$ gives a simplex-projection link. Negative Shannon entropy, $\Phi_T(q)=T\sum_jq_j\log q_j$, gives $G_{\Phi_T}(a)=T\log\sum_j\exp(a_j/T)$ and $L_{\Phi_T}(p,y)=-T\log p_y$. All $T\ge1/(2\beta)$ satisfy Definition~\ref{def:entropy}: for tangent vectors $v$, $\sum_jv_j^2/q_j\ge\|v\|_1^2\ge2\|v\|^2$. More generally, adding any continuous convex face-differentiable function to the quadratic entropy gives a member of the family. These examples illustrate reuse across different matching potentials, although the different temperatures of Shannon entropy alone merely rescale the canonical log loss.

\subsection{Interpreting the exact representation theorem}
\label{sec:existence}

The forecast is deterministic and uses at most $\min\{d,K-1\}$ independent
scores. To see this, factor $W=QA$ with $Q$ isometric onto the range of $W$;
then $\|A\|_F=\|W\|_F$ and $t\mapsto\rho(Qt)$ retains the same cap. No assumption is made on $Y\mid X$, no compactness of the entropy family is required, and the cap $\beta$ is independent of the distribution. We do not assert calibration or a convex-gradient representation of $\rho$. The theorem is an existence statement.

\subsection{Consequences of the exact lower bound}
\label{sec:sharp}
Theorem~\ref{thm:rank} quantifies both representation restrictions on
one quadratic instance. Use $d_\star=\min\{d,K-1\}$ and
$c=\beta R/\sqrt{d_\star}$ as in that theorem.

The first term is the cost of discarding directions; the second is the
cost of insufficient slope along retained directions. At $B=\beta$,
the second term vanishes and the formula becomes
\begin{equation}
 \frac{\beta R^2L^2(d_\star-r)}{2d_\star (d_\star+2)}.
 \label{eq:rank-regret}
\end{equation}
Thus $\min\{d,K-1\}$ scores are necessary in the worst case, even with an arbitrary measurable link. In particular, $K-1$ are necessary whenever $d\ge K-1$. At full rank $r=d_\star$, the regret is
$R^2L^2(\beta-B)_+^2/(2\beta(d_\star+2))$, showing quantitatively why the
original slope cannot be reduced while retaining the norm budget.
When fewer directions are retained, their larger available coefficient
norm offsets part of a slope reduction: the slope penalty vanishes
already at $B\ge\beta\sqrt{r/d_\star}$.

To prove the result, choose an isometry $U:\R^{d_\star}\to H$ and conditional
probabilities $\eta(x)=u+cUx$. A radius-$R$ comparator is Bayes-optimal,
and, for interior reports, quadratic regret is squared probability error divided by $2\beta$.
Reflection symmetry bounds the correlation of any learned output
coordinate with a feature direction by $B$ times the corresponding
singular value of its score matrix. Cauchy--Schwarz and Jensen's
inequality then give \eqref{eq:joint-main}. Equal singular values on
the retained subspace, followed by an affine probability map, attain
it. Appendix~\ref{sec:core-appendix} supplies the full argument,
including boundary reports and the unrestricted-head rank lower bound.

\paragraph{Why the probability link must be joint.}
The existence theorem uses all $K-1$ scores jointly. Requiring
$p_j(x)=\rho_j(a_j^\top x)$ for each $j<K$, with the last probability
obtained by subtraction, is a stronger restriction. For every $K\ge3$,
Theorem~\ref{separate:impossibility} constructs a spherical feature law and
a realizable softmax model on which even one probability constrained to a
single linear score incurs a uniform positive log-loss gap. This holds for
arbitrary measurable scalar links, so adding monotonicity cannot repair it.

The obstruction concerns separate \emph{probability} links. Scalar logits
followed by normalization already give a joint probability map, and are not
ruled out. It also differs from inner-product monotonicity of a joint link:
the Dirichlet construction in Section~\ref{sec:dirichlet-learning} retains that
property under its additional feature condition. Appendix~\ref{separate:section}
contains the full statement and proof.

\section{Our Approach and Techniques}
\label{sec:techniques}
We explain the exact upper bound first, then the matching lower bound.
The global-action upper bound rests on two inequalities involving the same residual.
One comes from fitting the probability link; the other comes from choosing
the linear scores. Strong convexity makes their error terms cancel.
The argument is easiest to see for an interior forecast. We explain
afterward why it remains valid on the boundary, including for log loss.

\subsection{The upper bound: one fit, all losses}
\label{sec:tech-upper}
\paragraph{Start by fixing the scores.}
Fix a coefficient matrix $W$ and fit a $\beta$-Lipschitz probability
link by least squares. Write
\[
 P=\rho_W(WX),\qquad \xi=P-e_Y,
 \qquad
 \rho_W\in\argmin_{\Lip(\rho)\le\beta}
              \E\|\rho(WX)-e_Y\|^2.
\]
The possible prediction vectors form a convex set. Thus the fitted
random vector is unique, although the link may remain undetermined
away from the score support. Its first-order optimality condition is
\begin{equation}
       \E\langle \xi,Q-P\rangle\ge0
       \qquad\text{for every other feasible prediction }Q.
       \label{eq:projection-main}
\end{equation}
The task is to choose useful feasible perturbations. A general change
to the link might violate its slope bound. Postcomposition by a
nonexpansive map does not: it can only decrease distances between
predictions. This elementary closure property is the source of the
simultaneous guarantee.

\paragraph{Turn convex functions into feasible tests.}
Let $F$ be a smooth convex function of the prediction. Its proximal map
is the minimizer
\[
 \prox_{tF}(z)=\argmin_v\left\{F(v)+\frac{\|v-z\|^2}{2t}\right\}.
\]
It moves a point a short distance in the direction $-\nabla F$ and is
nonexpansive. Apply this map to the fitted probabilities and project
back to the simplex. The resulting link remains feasible. Substituting
it in~\eqref{eq:projection-main}, dividing by the step size, and taking
the limit gives
\begin{equation}
                  \E\langle \xi,\nabla F(P)\rangle\le0.
                  \label{eq:convex-main}
\end{equation}
This is a one-sided statement: the residual has nonpositive average
correlation with every convex-gradient test. We do not require every
such correlation to vanish. In particular, the proof does not first
construct a calibrated or multicalibrated forecast.

Now fix any target entropy $\Phi\in\mathcal E_\beta$.
Its defining curvature assumption says that
\[
                  F(q)=\Phi(q)-\frac{\|q-u\|^2}{2\beta}
\]
is convex. Put $b=\E\langle \xi,P-u\rangle$.
Applying~\eqref{eq:convex-main} to this $F$ yields
\begin{equation}
  \E L_\Phi(P,Y)
     =-\E\Phi(P)+\E\langle \xi,\nabla\Phi(P)\rangle
     \le-\E\Phi(P)+\frac b\beta.
     \label{eq:forecast-main}
\end{equation}
The fitted link was chosen before $\Phi$. The curvature lower bound
is precisely what lets the same fit control all target entropies.

\paragraph{Use the fitted score once more.}
Convex tests control functions of the prediction, whereas the comparator
is a linear function of the features. We connect the two by interpolating
the fitted link with its own score:
\[
 s\longmapsto\Pi_{\Delta_K}
             \big((1-t)\rho_W(s)+t(u+\beta s)\big),
             \qquad 0\le t\le1.
\]
Both terms inside the convex combination have Lipschitz bound $\beta$,
and projection is nonexpansive. Hence this is another admissible
perturbation. Its first-order inequality is
\begin{equation}
                  \frac b\beta\le\E\langle \xi,WX\rangle.
                  \label{eq:own-main}
\end{equation}
The same number $b/\beta$ that upper-bounds the forecast's loss now
appears as a lower bound on a residual correlation with a linear score.

\paragraph{Choose scores that control every comparator.}
So far $W$ has been arbitrary. Define the continuous residual field
\[
                g(W)=\E[(P_W-e_Y)X^\top].
\]
We need a score matrix at which no feasible comparator decreases this
linear residual functional. For a compact convex coefficient set
$\mathcal W$, consider the map
\[
                W\longmapsto\Pi_{\mathcal W}^{F}(W-g(W)),
\]
where projection uses the Frobenius inner product. Brouwer's theorem
gives a fixed point. Projection optimality there says
\begin{equation}
 \E\langle \xi,VX\rangle\ge\E\langle \xi,WX\rangle
                       \ge b/\beta
                       \qquad(V\in\mathcal W).
 \label{eq:linear-main}
\end{equation}
No iteration convergence is asserted: this is the existence step.
Continuity follows because fitted predictions vary continuously in
$L^2$ with $W$; truncating $\|X\|$ then gives continuity of $g$ under
only a first-moment assumption.

Finally, the definition of the conjugate potential gives, for every
comparator $V$,
\begin{equation}
 \E\ell_\Phi(VX,Y)
       \ge-\E\Phi(P)+\E\langle \xi,VX\rangle
       \ge-\E\Phi(P)+b/\beta.
       \label{eq:comparator-main}
\end{equation}
Comparing~\eqref{eq:forecast-main} and~\eqref{eq:comparator-main}
proves exact omniprediction. We never need to know the sign of $b$.
The proof works because the \emph{same} residual term appears in the
two estimates with the same coefficient.

\begin{figure}[t]
\centering
\begin{tikzpicture}[font=\small,box/.style={draw=inkblue,rounded corners=2pt,
  fill=softblue,align=center,inner sep=8pt,text width=5.5cm},
  >=Latex]
\node[box] (fit) at (0,0) {Least-squares fit of the link\\
Nonexpansive changes preserve its slope};
\node[box] (scores) at (6.45,0) {Fixed point for the scores\\
Every comparator obeys the residual inequality};
\node[align=center,text width=5.7cm] (left) at (0,-1.6)
 {$\displaystyle \E L_\Phi(P,Y)\le -\E\Phi(P)+b/\beta$};
\node[align=center,text width=5.7cm] (right) at (6.45,-1.6)
 {$\displaystyle \E\ell_\Phi(VX,Y)\ge-\E\Phi(P)+b/\beta$};
\draw[->,thick,inkblue] (fit.south) -- (left.north);
\draw[->,thick,inkblue] (scores.south) -- (right.north);
\node[draw=inkblue,fill=softgray,rounded corners=2pt,align=center,
  text width=8cm,inner sep=7pt] (end) at (3.225,-3.05)
 {One forecast competes with every comparator\\for every admissible entropy};
\draw[->,thick,inkblue] (left.south) -- (end.north west);
\draw[->,thick,inkblue] (right.south) -- (end.north east);
\end{tikzpicture}
\caption{The two parts of the exact certificate. Both inequalities contain
the same residual term $b/\beta$, so no approximation loss remains.
For a Dirichlet dictionary, the least-squares fit supplies the
right-hand inequality directly.}
\label{fig:cancellation}
\end{figure}
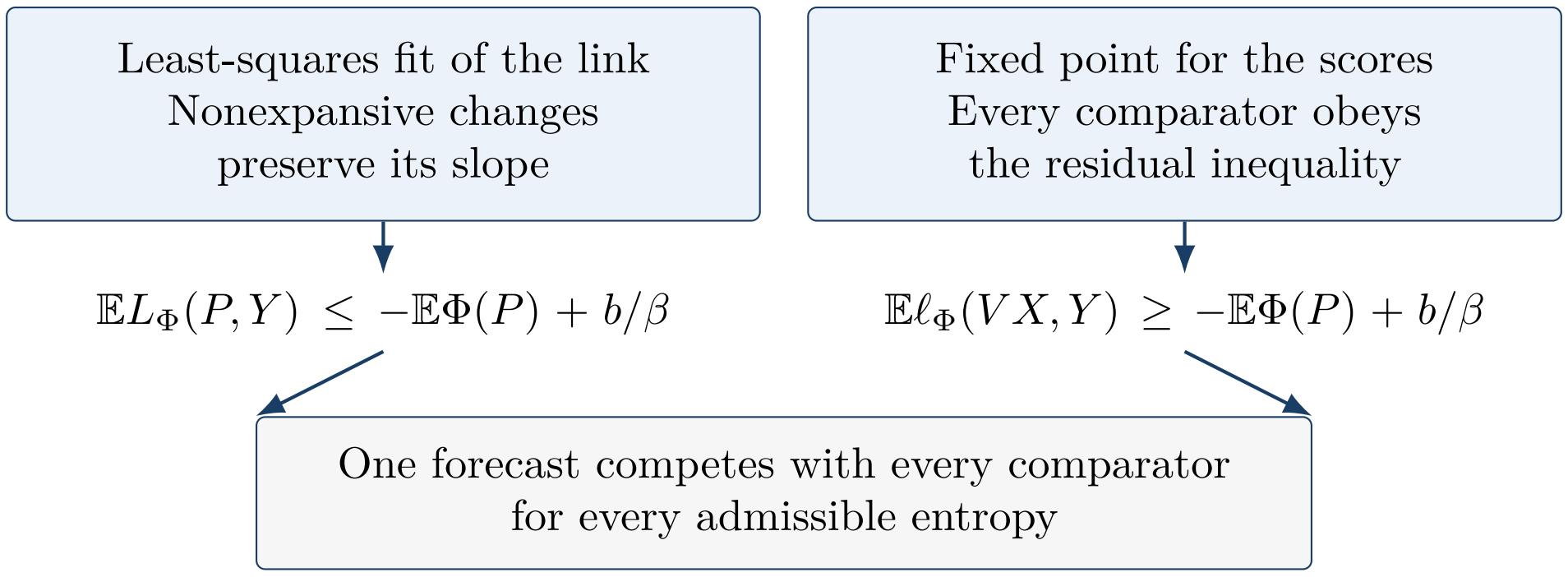

\paragraph{Why boundary probabilities cause no loss.}
A fitted probability can be zero, and an entropy gradient can diverge
there. Simply differentiating through projection would therefore leave
a gap. We first project the fitted probabilities onto the trimmed simplex
$\{q:q_j\ge t\}$. This is a feasible nonexpansive transformation.
If the true label had zero fitted probability, the first-order change
in squared loss would be strictly negative. Optimality rules this out:
$P_j=0$ implies $Y\ne j$ almost surely. The residual is consequently
tangent to the smallest face containing $P$.

This tangency makes projection invisible in the residual inner products
used above. To handle an entropy with unbounded derivatives, we first
test a smooth convex approximation on a trimmed simplex and then pass
to its relative derivative on each face. A supporting-hyperplane lower
bound permits Fatou's lemma and also proves integrability. Thus the
boundary argument establishes the exact theorem, including log loss.
For implementation, the decision rule may instead evaluate the entropy
gradient at $(1-t)P+tu$. For each target, its error tends to zero as
$t$ tends to zero; the forecast itself need not be relearned.

\paragraph{Why the binary link remains monotone.}
In one dimension a convex proximal map is nondecreasing. Clipping and
translations followed by clipping are also nondecreasing, and the
own-score interpolation combines two nondecreasing functions. The
entire proof therefore stays inside monotone single-index models.
In several dimensions these facts give a joint Lipschitz map, but they
do not give cyclic monotonicity. Keeping this distinction explicit is
essential to the multiclass statement. With bounded binary actions,
least squares and the fixed point first give a sequence of increasingly
accurate one-score forecasts. A compactness argument using the one-dimensional Bayes comparison
then gives one exact forecast (Appendix~\ref{sec:bounded-binary-proof}).

\paragraph{From existence to sample bounds.}
The certificate also measures approximate solutions. On a Frobenius
ball define
\[
          \Gamma(W)=\langle g(W),W\rangle_F+R\|g(W)\|_F.
\]
Every target's regret is at most $\Gamma(W)$. The exact proof retains
a further negative squared-distance term
(Lemma~\ref{core:certificate}); in particular a zero-gap fit recovers
any realizable model in the comparator class. A finite matrix net
therefore yields a statistical learner once the empirical risks can
be controlled uniformly. This search need not be efficient.

\subsection{The lower bound: lost directions and insufficient slope}
\label{sec:tech-lower}
The lower bound uses a feature distribution with no distinguished
direction. Let $d_\star=\min\{d,K-1\}$, take $X$ uniformly from a small
Euclidean ball in $\R^{d_\star}$, and let
\[
            \eta(x)=u+cUx,
            \qquad c=\beta R/\sqrt{d_\star},
\]
where $U$ is an isometric embedding into $H$.
Choose the ball small enough that every coordinate of $\eta(x)$ is
strictly positive. For the quadratic entropy, a norm-$R$ linear action
realizes these conditional probabilities exactly. Its risk is the
Bayes risk. For an interior forecast, the regret is its squared
distance from $\eta$, divided by $2\beta$; the same expression
lower-bounds the regret of every forecast.

\paragraph{A missing score loses an entire direction.}
Suppose a forecast depends on linear scores with row space $E$ of
dimension $r$. Conditional on those scores, reflection symmetry gives
$\E[X\mid WX]=\Pi_E X$. Even an arbitrary measurable link cannot
recover the component perpendicular to $E$. Its best possible prediction
is the conditional mean $u+cU\Pi_E X$, and its irreducible squared
error is
\[
              c^2\E\|X-\Pi_E X\|^2
                 =\frac{L^2}{d_\star+2}(d_\star-r)c^2.
\]
This proves necessity of the score count without any regularity
assumption on the link. It also explains why a scalar probability
output cannot evade the lower bound by storing a complicated function:
the information was lost before the link was applied.

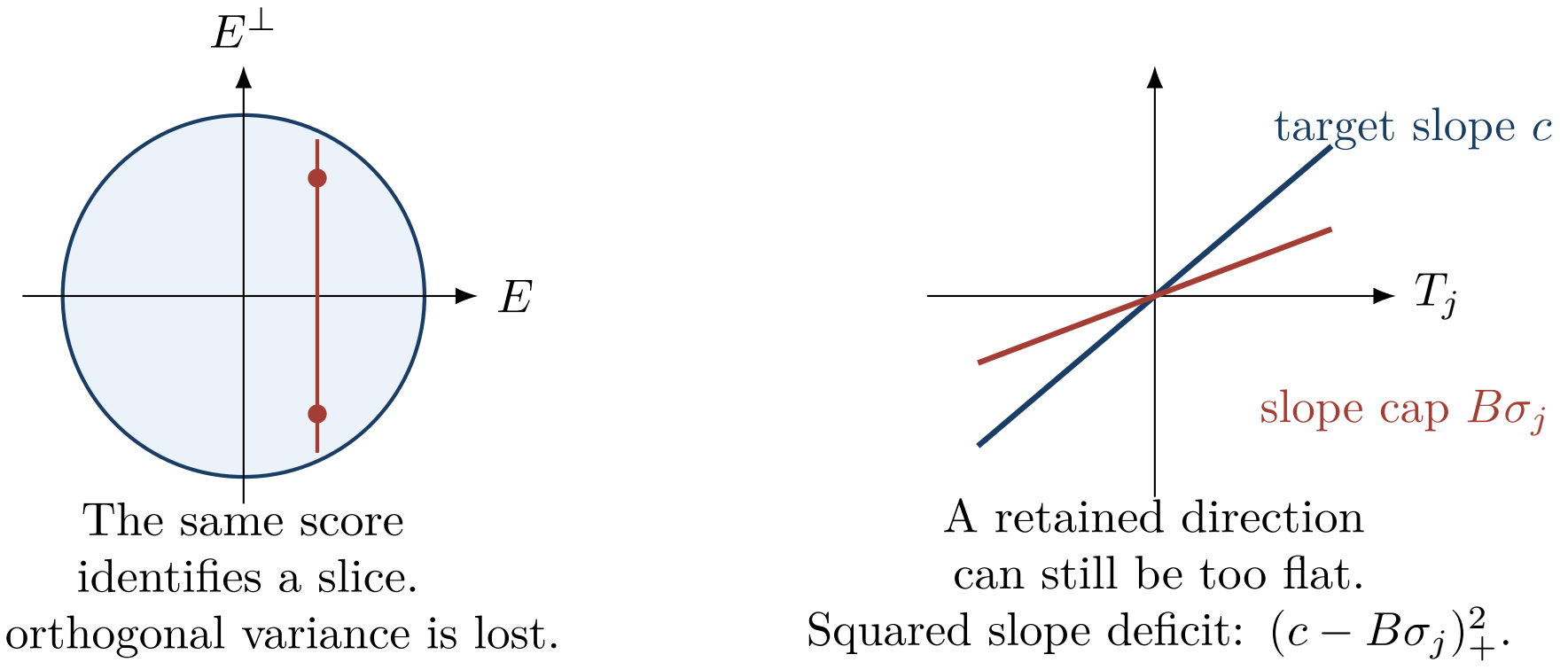
\begin{figure}[t]
\centering
\begin{tikzpicture}[font=\small,>=Latex]
\begin{scope}[shift={(0,0)}]
\draw[fill=softblue,draw=inkblue,thick] (0,0) circle (1.35);
\draw[->] (-1.65,0)--(1.75,0) node[right] {$E$};
\draw[->] (0,-1.55)--(0,1.72) node[above] {$E^\perp$};
\draw[warmred,thick] (0.55,-1.17)--(0.55,1.17);
\fill[warmred] (0.55,0.88) circle(2pt);
\fill[warmred] (0.55,-0.88) circle(2pt);
\draw[densely dashed,warmred] (0.55,0.88)--(0.55,0);
\draw[densely dashed,warmred] (0.55,-0.88)--(0.55,0);
\node[align=center,text width=4.6cm] at (0,-2.12)
 {The same score identifies a slice.\\Its orthogonal variance is lost.};
\end{scope}
\begin{scope}[shift={(6.8,0)}]
\draw[->] (-1.7,0)--(1.8,0) node[right] {$T_j$};
\draw[->] (0,-1.5)--(0,1.72);
\draw[inkblue,very thick] (-1.32,-1.12)--(1.32,1.12);
\draw[warmred,very thick] (-1.32,-0.5)--(1.32,0.5);
\node[anchor=west,inkblue] at (0.75,1.25) {target slope $c$};
\node[anchor=west,warmred] at (0.65,-0.87) {slope cap $B\sigma_j$};
\node[align=center,text width=5.2cm] at (0,-2.12)
 {A retained direction can still be too flat.\\Squared slope deficit: $(c-B\sigma_j)_+^2$.};
\end{scope}
\end{tikzpicture}
\caption{The two terms in the exact lower bound. The left panel shows a
two-dimensional section of the hard instance. The right panel illustrates
the slope constraint along one retained singular direction when its
allowed slope is smaller than the target slope.}
\label{fig:rank-slope}
\end{figure}

\paragraph{A small slope leaves error in the retained directions.}
Write $\sigma_1,\ldots,\sigma_r$ for the nonzero singular values of
the score matrix. Along its $j$th right singular direction, a
$B$-Lipschitz link changes the prediction at speed at most
$B\sigma_j$. Reflecting that coordinate bounds its correlation with
the target by $B\sigma_j\E X_j^2$. Cauchy--Schwarz then gives
squared error at least
\[
                   \E X_j^2(c-B\sigma_j)_+^2.
\]
The coefficient budget says $\sum_j\sigma_j^2\le R^2$.
The function $t\mapsto(c-B\sqrt t)_+^2$ is convex and nonincreasing,
so Jensen's inequality shows that equal singular values minimize
the total error. Setting $\sigma_j=R/\sqrt r$ gives the formula
in Theorem~\ref{thm:rank}.

Every step is attained: keep any $r$ directions, give them equal
coefficient norm, and use slope
$\min\{c,BR/\sqrt r\}$ on that subspace. The probabilities stay
inside the simplex on the feature support; projection defines the
same-cap link everywhere else. Hence the formula is an exact tradeoff,
not just an asymptotic obstruction. The two restrictions also interact:
discarding directions frees coefficient norm for those retained, which
explains the threshold $B=\beta\sqrt{r/d_\star}$ at which the additional
slope penalty disappears.

\section{Statistical Learning with Few Scores}
\label{sec:statistics}
We prove Theorem~\ref{thm:pac} by an empirical certificate and uniform
convergence. Throughout this section, $\|X\|\le L$, $A=LR$, and the
smooth family $\mathcal E(\beta,\lambda,M)$ and finite action
$k_\Phi=\nabla_H\Phi$ are as in Section~\ref{sec:main-results}.

The learned link retains the original cap $\beta$. We use a finite quadratic-program extension, described below, to interpolate the fitted probabilities without increasing their Lipschitz constant. For $K=2,3$, \eqref{eq:sample-main} is quadratic in inverse accuracy, up to logarithms. For $K\ge4$, no optimality of the larger exponent is claimed. A quadratic-entropy lower bound, already under spherical feature marginals, matches the accuracy exponent for $K=2,3$ when ambient dimension may grow (Proposition~\ref{stat:lower}). In the scalar binary model, monotone fitting and piecewise-linear interpolation retain a nondecreasing $\beta$-Lipschitz head, with sample size $\widetilde O((M+LR+\beta\lambda LR)^2/\varepsilon^2)$ (Corollary~\ref{stat:binary}).

\paragraph{Certificate and statistical transfer.}
On the sample, fit $\beta$-Lipschitz probabilities at each score matrix. With $g_n(W)=\E_n[(P_W-e_Y)X^\top]$, define
\begin{equation}
 \Gamma_n(W)=\ip{g_n(W)}{W}_F+R\|g_n(W)\|_F.
 \label{eq:gap-main}
\end{equation}
The notation $\ip{\cdot}{\cdot}_F$ denotes the Frobenius inner product. The empirical counterpart of the cancellation proof bounds every target's empirical regret by $\Gamma_n(W)$. A fixed point has zero gap, and stability of the fit ensures that a finite matrix net contains a small-gap candidate. An explicit Lipschitz extension then defines its prediction on unseen features.

The extension is constructive. Given the fitted score--probability pairs, Lemma~\ref{stat:same-cap-extension} defines $\rho(s)$ by a finite convex quadratic program. It interpolates the data, takes values in their probability convex hull, and retains cap $\beta$. An objective error $\zeta$ in this query program changes the output by at most $\sqrt\zeta$; the exact program defines the globally Lipschitz head.

The statistical step controls the resulting family of scalar loss profiles after linear projection. Smoothness in \eqref{eq:smooth-family} makes each canonical loss $\sqrt2\lambda$-Lipschitz as a function of the reported probability. Entropy bounds for Lipschitz functions on an $m$-dimensional score ball, together with norm-controlled linear classes, give \eqref{eq:sample-main}. Appendix~\ref{sec:statistics-appendix} proves the covering, chaining, finite-search, and extension claims. The independence from $d$ concerns sample complexity; it does not make the unrestricted matrix search efficient. Singular entropies admit a separate extension: smoothing one learned forecast gives a joint guarantee for the smooth family and a bounded temperature range of log losses, with finite logits and the same norm and slope budgets. Corollary~\ref{stat:joint-log-pac} gives the slower $\widetilde O_K(\varepsilon^{-4})$ rate for $K=2,3$.

The lower bound uses just one quadratic entropy and a uniform spherical
feature marginal. This example belongs to the general bounded-feature
class and gives a lower bound even for learners that can return an
arbitrary probability forecast. A similar construction for one-hot
features gives the dictionary lower bound below.

\section{Sample Complexity on Dirichlet Dictionaries}
\label{sec:dirichlet-learning}
The Dirichlet feature condition lets us obtain the quadratic rate in
Theorem~\ref{thm:dirichlet-learning} by fitting an affine map to the
labels and projecting its output onto the simplex.

Let the feature law be supported on a fixed finite dictionary
$x_1,\ldots,x_N\in\R^d$, with $\|x_i\|\le L$. Let $\mathsf X$
have these vectors as rows, and define
\[
 J_N=I_N-\frac1N\mathbf1\mathbf1^\top,\qquad
 G=J_N\mathsf X\mathsf X^\top J_N.
\]
The \emph{Dirichlet feature condition} is
\begin{equation}
 \operatorname{rank}G=N-1,\qquad
 (G^\dagger)_{ij}\le0\quad(i\ne j),
 \label{eq:dirichlet-condition}
\end{equation}
where $G^\dagger$ is the Moore--Penrose pseudoinverse. For $N=1$,
use $G^\dagger=0$. The condition makes the minimum affine
interpolation norm a graph Dirichlet energy. The rank condition requires affine independence, hence
$N\le d+1$. Scaled one-hot features $x_i=Le_i$ satisfy it because
$G^\dagger=L^{-2}J_N$. The feature probabilities and conditional
label distributions are arbitrary.

Use the smooth entropy family $\mathcal E(\beta,\lambda,M)$ from
Section~\ref{sec:main-results}, with the bounds in \eqref{eq:smooth-family},
and the same finite canonical action $k_\Phi(q)=\nabla_H\Phi(q)$ at all
probabilities, including the boundary.

The learner minimizes the convex raw-score objective
\begin{equation}
 \min_{a\in H,\,\|W\|_F\le R}
 \frac1{2n}\sum_{i=1}^n
       \|a+\beta W X_i-(e_{Y_i}-u)\|^2
 \label{eq:dirichlet-program}
\end{equation}
and projects onto the simplex when predicting. Among minimizers it
chooses the weight matrix of smallest Frobenius norm. This choice
determines the optimal intercept, which belongs to the learned link. The feature condition makes nonexpansive
postprocessing preserve the feasible prediction set, so an optimizer
satisfies the simultaneous empirical certificate. The constant in the sample
bound is independent of $K$ at fixed $M,\lambda,L,R,\beta$.
Appendix~\ref{app:dirichlet} proves the certificate and transfers it to
population risk.

\paragraph{Optimal sample exponent.}
The exponent two is necessary for every fixed $K$, already for the
quadratic entropy and a uniform one-hot dictionary. In that construction,
the ambient dimension and dictionary size grow as
$\Theta_K(1/\varepsilon)$. Proposition~\ref{prop:dirichlet-lower}
proves an $\Omega_K(\varepsilon^{-2})$ lower bound even for unrestricted
output forecasts. Thus the worst-case accuracy exponent over these
dictionaries matches the exponent of the upper bound.

\section{Further Questions}
Can the same
architecture be learned in polynomial time under an arbitrary feature
marginal? The finite search establishes statistical learnability, but its
running time may be exponential. Can the general multiclass sample bound
be made quadratic in inverse accuracy for every fixed $K$? The Dirichlet
theorem attains this rate under a specific geometric assumption; for
unrestricted features and $K\ge4$, the optimal exponent remains open.
For $K\ge3$, can a forecast with the same score count also achieve exact
omniprediction with bounded downstream actions? In the global-action
model, can the joint link always be chosen as the gradient of a convex
function?

\vspace{1em}
\noindent\begin{minipage}{\textwidth}
\textbf{AI Use Statement.}\quad
This work resulted from extensive interaction between the authors and OpenAI
Astra, with the authors providing suggestions and critically checking Astra's
responses through iterative exchanges. OpenAI Astra was also used in writing
and revising the manuscript through detailed prompts from the authors, with
Gemini Argon assisting in some parts. The authors directed the process and
evaluated the proposed ideas and arguments. The accompanying Lean~4 files are available in the
\href{https://github.com/ZHadizadeh/SIM-Omniprediction}{project repository}.
\par
\end{minipage}

\begingroup
\small
\setlength{\bibsep}{3pt}

\endgroup
\clearpage
\appendix

\section{Binary monotone single-index existence}
\label{sec:binary-appendix}
\subsection{Proof for global actions}
We prove Theorem~\ref{thm:binary} in the scalar convention of Section~\ref{sec:binary}. Let $\mathcal W\subset\R^d$ be any nonempty compact convex set; the radius-$R$ Euclidean ball is the main special case. For each $w$, minimize $\E(\rho(w^\top X)-Y)^2$ over nondecreasing $\beta$-Lipschitz maps $\rho:\R\to[0,1]$. Write $P_w$ for the fitted random variable. The profile class is convex and sequentially compact for locally uniform convergence. Dominated convergence gives a minimizer; strict convexity of squared loss gives uniqueness of $P_w$ as a random variable. If $w_n\to w$, take a locally uniformly converging subsequence of optimal profiles. Their predictions converge almost surely and in $L^2$ to a feasible prediction at $w$. Comparing their risks with each fixed feasible profile proves optimality of the limit. Uniqueness then gives $P_{w_n}\to P_w$ in $L^2$ along the whole sequence.

Put $\xi=P_w-Y$. For any feasible prediction $Q$, projection optimality gives
\begin{equation}
 \E[\xi(Q-P_w)]\ge0.\label{eq:binary-projection}
\end{equation}
The clipped profile $\Pi_{[t,1-t]}\circ\rho_w$ is feasible for $0<t<1/2$. Its derivative at $t=0$ is $1$ on $\{P_w=0\}$, $-1$ on $\{P_w=1\}$, and zero elsewhere. Therefore \eqref{eq:binary-projection} forces
\[
 \E\big[Y\mathbf1_{\{P_w=0\}}+(1-Y)\mathbf1_{\{P_w=1\}}\big]=0.
\]
Both terms are nonnegative, so $Y=0$ on $\{P_w=0\}$ and $Y=1$ on $\{P_w=1\}$ almost surely. In particular $\xi=0$ at the two endpoints. This makes clipping invisible in residual inner products to first order. Applying \eqref{eq:binary-projection} to $\Pi_{[0,1]}(P_w\pm t)$ gives $\E \xi=0$.

Let $F:\R\to\R$ be finite, convex, differentiable, and have a globally Lipschitz derivative. Its proximal map $T_t=\operatorname{prox}_{tF}$ is nondecreasing and $1$-Lipschitz: for $a=T_t(x)$ and $b=T_t(y)$, monotonicity of $F'$ in the equations $x=a+tF'(a)$ and $y=b+tF'(b)$ gives $0\le a-b\le x-y$ when $x\ge y$. Thus $\Pi_{[0,1]}\circ T_t\circ\rho_w$ remains a nondecreasing $\beta$-Lipschitz profile. The proximal equation gives $(T_t(p)-p)/t\to-F'(p)$ uniformly for $p\in[0,1]$. Divide \eqref{eq:binary-projection} by $t$ and use $\xi=0$ at endpoints to conclude
\begin{equation}
 \E[\xi F'(P_w)]\le0.\label{eq:binary-convex}
\end{equation}
This inequality extends to every finite continuous convex $F$ on $[0,1]$ differentiable in its interior, with endpoint derivatives interpreted as zero in products with $\xi$. Here are the details of the extension. For fixed $s\in(0,1/2)$, let $F_\delta$ be the Moreau envelope of the extension of $F$ by $+\infty$ outside $[0,1]$. Apply the smooth test to $p\mapsto F_\delta((1-2s)p+s)/(1-2s)$. As $\delta\downarrow0$, its derivative converges uniformly for $p\in[0,1]$ to $F'((1-2s)p+s)$, because the arguments lie in $[s,1-s]$. Thus \eqref{eq:binary-convex} holds with $F'((1-2s)P_w+s)$. As $s\downarrow0$, the residual products converge pointwise on the interior, and they vanish at the endpoints. More explicitly, if $\operatorname{osc}(F)=\max_{[0,1]}F-\min_{[0,1]}F$, the supporting inequalities at $Y$ and $1/2$ give
\[
 (P_w-Y)F'((1-2s)P_w+s)
 \ge -\frac{\operatorname{osc}(F)}{1-2s}\ge-2\operatorname{osc}(F),\qquad 0<s\le1/4.
\]
Fatou's lemma after adding $2\operatorname{osc}(F)$ yields the assertion, including integrability of the limiting residual product.

The map $a\mapsto\Pi_{[0,1]}((1-t)\rho_w(a)+t\beta a)$ is nondecreasing and $\beta$-Lipschitz for $0\le t\le1$. Its projection inequality gives, by dominated convergence using $\E\|X\|<\infty$,
\begin{equation}
 \beta\E[\xi w^\top X]\ge\E[\xi P_w].\label{eq:binary-own}
\end{equation}
The vector field $g(w)=\E[(P_w-Y)X]$ is continuous. To prove this with only a first moment, truncate $\|X\|$; on the bounded part use $L^2$ continuity of $P_w$, while the tail is bounded uniformly by $\E[\|X\|\mathbf1_{\{\|X\|>T\}}]$. Brouwer's theorem applied to $w\mapsto\Pi_{\mathcal W}(w-g(w))$ gives a fixed point. At that point, with $P=P_w$ and $b=\E[(P-Y)P]$,
\[
 \sup_{v\in\mathcal W}\E[(Y-P)v^\top X]
 \le-\E[(P-Y)w^\top X]\le-b/\beta.
\]

For every target $\varphi$, apply the extended convex-test inequality to $F(p)=\varphi(p)-p^2/(2\beta)$. Endpoint residuals vanish, so
\[
 \E L_\varphi(P,Y)
 =-\E\varphi(P)+\E[(P-Y)\varphi'(P)]
 \le-\E\varphi(P)+b/\beta,
\]
where the middle expression uses the stipulated face convention at endpoints. Fenchel's inequality gives for every $v\in\mathcal W$
\[
 \E\ell_\varphi(v^\top X,Y)
 \ge-\E\varphi(P)+\E[(P-Y)v^\top X]
 \ge-\E\varphi(P)+b/\beta.
\]
The forecast risk is finite: the canonical loss is bounded below by $-\varphi(Y)$, and the upper bound above is finite. Comparator risks are finite because $g_\varphi$ is $1$-Lipschitz and $X$ is integrable. The fixed point was chosen independently of $\varphi$, proving simultaneous exact comparison and Theorem~\ref{thm:binary}.

\paragraph{Why monotonicity is preserved.}
Every perturbation used above is a nondecreasing postcomposition, except the own-score interpolation, which is directly a convex combination of two nondecreasing functions followed by clipping. Thus no step optimizes over a larger class of nonmonotone scalar profiles. The vector theorem, by contrast, permits all Lipschitz simplex-valued profiles; its proof does not establish cyclic monotonicity of the learned link.

\subsection{Proof of the bounded-action theorem}\label{sec:bounded-binary-proof}

\begin{proof}[Proof of Theorem~\ref{thm:bounded-binary}]
Write $\mathcal W=\{v:\|v\|_2\le R\}$. We first construct approximate
forecasts with a freely chosen slope bound $B>0$. Apply the least-squares
fit and projected fixed point used in the proof of Theorem~\ref{thm:binary},
with $B$ in place of $\beta$. Those steps use only $0\le Y\le1$:
in particular, the two terms in the endpoint clipping argument remain
nonnegative. For the resulting $P_B=\rho_B(w_B^\top X)$, put
$\xi_B=P_B-Y$. We have $\E \xi_B=0$, and \eqref{eq:binary-own} and the
fixed-point inequality give
\[
 \sup_{v\in\mathcal W}\E[(Y-P_B)v^\top X]
 =\E[(Y-P_B)w_B^\top X]\le -\E[\xi_BP_B]/B\le 1/(16B).
\]
Indeed $\E[\xi_BP_B]=\operatorname{Var}(P_B)-\operatorname{Cov}(Y,P_B)
\ge s^2-s/2\ge-1/16$, where $s^2=\operatorname{Var}(P_B)$ and
$\operatorname{Var}(Y)\le1/4$.

We also have $\E[\xi_B f(P_B)]\le0$ for every bounded nondecreasing
$f:[0,1]\to\R$. For Lipschitz $f$ this follows from
\eqref{eq:binary-convex} by taking a primitive of $f$, extended
constantly beyond the endpoints before integration. For general $f$,
interpolate its values linearly on successively finer grids containing
the endpoints and increasingly many of its countably many jump points.
These bounded nondecreasing interpolants converge pointwise, including
at each jump, so dominated convergence gives the claim.

For integer $B\to\infty$, pass to a weakly converging subsequence of
the laws of $(X,Y,P_B)$, supported on a compact set. Denote the limit
by $(X,Y,Q)$, with its original $(X,Y)$ marginal. Testing first with
continuous functions and then using the same interpolation for $f$ yields
\[
 \E[(Y-Q)v^\top X]=0\quad(v\in\mathcal W),\qquad
 \E[(Q-Y)f(Q)]\le0\quad(f\text{ bounded and nondecreasing}).
\]
The first equality uses the symmetry of $\mathcal W$.
Let $H_\sigma(q)=\max_{t\in I}\{qt-G_\sigma(t)\}$. For $U=P_B$ or $Q$,
Bayes optimality and the second test with $k_\sigma$ give
\[
 \E\ell_\sigma(k_\sigma(U),Y)
 =-\E H_\sigma(U)+\E[(U-Y)k_\sigma(U)]\le-\E H_\sigma(U).
\]
Define $\mathcal G(\sigma,v)=\E\ell_\sigma(v^\top X,Y)+\E H_\sigma(Q)$.
Moment equality makes this the expectation of the nonnegative quantity
$G_\sigma(v^\top X)-Qv^\top X+H_\sigma(Q)$. Thus $Q$ attains the
desired comparison for all $\sigma,v$. It remains to obtain a
deterministic one-score forecast.

The continuous function $\mathcal G$ attains its minimum $\gamma\ge0$ on
$\mathcal S\times\mathcal W$. Indeed
$\|G_\sigma-G_\tau\|_\infty,\|H_\sigma-H_\tau\|_\infty
\le A\|\sigma-\tau\|_\infty$, and the action losses are uniformly
Lipschitz on $I$. If $\gamma>0$, the $H_\sigma$ form a compact family of
continuous functions on $[0,1]$. A finite uniform net and weak
convergence give $\sup_\sigma|\E H_\sigma(P_B)-\E H_\sigma(Q)|<\gamma$
for some $B$ in the subsequence. The preceding risk bound then puts
$P_B$ below every comparator risk.

If $\gamma=0$, take a minimizing pair $(\sigma_*,v_*)$.
Its nonnegative integrand vanishes almost surely, so $t=v_*^\top X$
maximizes $Qt-G_{\sigma_*}(t)$ on $I$. In the interior this gives
$Q=\sigma_*(t)$. At $t=-A$ it gives $Q\le\sigma_*(-A)$, and at
$t=A$ it gives $Q\ge\sigma_*(A)$. Equality in Cauchy--Schwarz
implies that each endpoint score has at most one feature preimage
in the radius-$L$ ball. At any such point $x$ of positive mass,
pass to a further subsequence on which the deterministic predictions
$\rho_B(w_B^\top x)$ converge to $c_x$. The limit joint law of $(X,Q)$
contains an atom of mass at least $\Pr(X=x)$ at $(x,c_x)$: subtract
the corresponding atom from each prelimit law and pass to the limit
against nonnegative continuous functions. The fixed $X$ marginal
then forces $Q=c_x$ on $\{X=x\}$. Use $\sigma_*$ on the interior of
$I$, the value $c_x$ at each endpoint of positive mass, and the
endpoint value of $\sigma_*$ otherwise. Constant extension off $I$
gives a Borel nondecreasing link by the endpoint inequalities.
Consequently $Q=\rho(v_*^\top X)$ almost surely, which completes the proof.
\end{proof}

\section{Proofs for global multiclass existence and sharpness}
\label{core:appendix}\label{sec:core-appendix}

We use the notation of Section~\ref{sec:model}. Put $\mathcal C=\Delta_K-u$.
For a nonempty support $J\subseteq[K]$, let
$H_J=\{v\in H:v_j=0\text{ for }j\notin J\}$.
The relative gradient $\nabla_J\Phi(p)$ belongs to $H_J$ and is zero
on singleton faces. We write $J(p)=\{j:p_j>0\}$ and use the
canonical boundary loss~\eqref{eq:extended}, also denoted by
\begin{equation}
 L_\Phi(p,y)=
 \begin{cases}
 -\Phi(p)+\langle\nabla_{J(p)}\Phi(p),p-e_y\rangle,&y\in J(p),\\
 +\infty,&y\notin J(p).
 \end{cases}
 \label{core:canonical}
\end{equation}
The support conclusion below ensures that the constructed forecast
never excludes its realized label.

\subsection{The canonical boundary loss}

\begin{lemma}[Boundary interpretation and properness]
\label{core:boundary}
Let $K\ge2$, and let $\Phi:\Delta_K\to\R$ be finite, continuous and
convex, and differentiable on the relative interior of every nonempty
face. The loss in \eqref{core:canonical} satisfies
$L_\Phi(p,y)\ge-\Phi(e_y)$ and is proper: for all $p,q\in\Delta_K$,
\[
 \sum_{y\in[K]}q_yL_\Phi(q,y)=-\Phi(q)
 \le\sum_{y\in[K]}q_yL_\Phi(p,y),
\]
with $0\cdot(+\infty)=0$ in these finite sums.
For each $p\in\Delta_K$ there is a sequence $(a_t)_{t\in\mathbb N}$
in $H$ such that $\ell_\Phi(a_t,y)\to L_\Phi(p,y)$ in
$(-\infty,+\infty]$ for every label $y$.
If $p\in\operatorname{relint}\Delta_K$, the finite action
$a=\nabla_H\Phi(p)$ attains these losses exactly.
\end{lemma}
\begin{proof}
Let $J=J(p)$ and $g=\nabla_J\Phi(p)$. For $y\in J$, convexity on
$\Delta_J$ gives
\[
 \Phi(e_y)\ge\Phi(p)+\langle g,e_y-p\rangle,
\]
which proves the lower bound; it is automatic for $y\notin J$.
For a true distribution $q$ with support contained in $J$,
\[
 \mathbb E_{Y\sim q}L_\Phi(p,Y)
 =-\Phi(p)+\langle g,p-q\rangle\ge-\Phi(q).
\]
If the support of $q$ is not contained in $J$, the left side is infinite.
Reporting $p=q$ attains $-\Phi(q)$, proving properness.

For $t>0$, put $\widetilde a_{t,j}=g_j$ when $j\in J$ and $\widetilde a_{t,j}=-t$
otherwise, and set
$a_t=\widetilde a_t-K^{-1}\langle \widetilde a_t,\mathbf1\rangle\mathbf1$.
Centering cancels in the loss, so
\[
 \ell_\Phi(a_t,y)
 =\max_{q\in\Delta_K}\{\langle \widetilde a_t,q\rangle-\Phi(q)\}-\widetilde a_{t,y}.
\]
The maximum converges to
$\max_{q\in\Delta_J}\{\langle g,q\rangle-\Phi(q)\}
=\langle g,p\rangle-\Phi(p)$.
To verify convergence, this face maximum is a lower bound for all $t$.
The quantities $\langle g,q_J\rangle-\Phi(q)$ are uniformly bounded,
so every maximizing $q_t$ has
$\sum_{j\notin J}q_{t,j}\to0$. Along each convergent subsequence,
continuity of $\Phi$ and the nonpositive penalty give an upper limit
at most the face maximum. Compactness supplies such subsequences and
proves convergence. Subtracting $\widetilde a_{t,y}$ gives the asserted finite
limit for $y\in J$ and divergence to $+\infty$ otherwise.
If $J=[K]$, the relative supporting inequality directly makes $p$ a
maximizer at the score $g=\nabla_H\Phi(p)$.
\end{proof}

Strong convexity also shows that the maximizer $q(a)$ in $G_\Phi(a)$ is
unique and satisfies
\[
 \|q(a)-q(b)\|\le\beta\|a-b\|.
\]
Indeed, adding the two strong-convexity optimality inequalities gives
$\beta^{-1}\|q(a)-q(b)\|^2
\le\langle a-b,q(a)-q(b)\rangle$.
Danskin's theorem then gives
$\nabla G_\Phi(a)=q(a)-u$.
Since $\mathcal C$ is bounded, $G_\Phi$ is globally Lipschitz, irrespective of
whether $\nabla_H\Phi$ diverges at the boundary.

\subsection{Lipschitz least-squares fits}

Fix $B>0$, a measurable $H$-valued score $S$, and a measurable
label $Y\in[K]$. A fit is $P=\rho(S)$ minimizing
$\mathbb E\|\rho(S)-e_Y\|^2$ over all $B$-Lipschitz maps
$\rho:H\to\Delta_K$. For a linear map $W:\R^d\to H$ and measurable
$X$, write $P_W$ for the fitted random prediction at $S=WX$.

\begin{lemma}[Existence and continuity of the fitted prediction]
\label{core:fit-continuity}
For every linear $W:\R^d\to H$, a fit $P_W$ exists and is unique as
a random prediction up to almost-sure equality.
If $W_n\to W$ in matrix norm, the corresponding fitted predictions
satisfy $P_{W_n}\to P_W$ in $L^2$.
If $\mathbb E\|X\|<\infty$, the matrix
\[
 g(W)=\mathbb E[(P_W-e_Y)X^\top]
\]
is well defined: its matrix-valued integrand is integrable, it maps
$\R^d$ into $H$, and it depends continuously on $W$.
\end{lemma}
\begin{proof}
The profile class is convex. It is sequentially compact for locally
uniform convergence: on each closed ball in the finite-dimensional
space $H$, its members are uniformly bounded and equicontinuous;
Arzel\`a--Ascoli followed by a diagonal subsequence yields a global
$B$-Lipschitz limit with values in the closed simplex.
Along this subsequence, predictions at $S$ converge pointwise and are
bounded, so their squared risks converge by dominated convergence.
This proves attainment. Strict convexity of squared distance on the
convex set of attainable prediction random variables proves uniqueness.

For continuity, take any subsequence of profiles fitted at $W_n$ and
extract a locally uniformly convergent further subsequence with limit
$\rho$. For every fixed $x$,
\[
 \|\rho_n(W_nx)-\rho(Wx)\|
 \le B\|(W_n-W)x\|+\|\rho_n(Wx)-\rho(Wx)\|\longrightarrow0.
\]
Boundedness gives $L^2$ convergence. Comparing the optimal risk at
$W_n$ with the risk of each fixed feasible profile and passing to the
limit shows that $\rho(WX)$ is optimal at $W$. Uniqueness identifies
it with $P_W$. Every subsequence has this property, establishing the
claimed convergence of the whole sequence.
Finally,
\[
 \|g(W_n)-g(W)\|_F
 \le\mathbb E[\|P_{W_n}-P_W\|\|X\|].
\]
For $T>0$, the contribution from $\|X\|\le T$ tends to zero by
$L^2$ convergence. The remainder is at most
$\sqrt2\,\mathbb E[\|X\|\mathbf1_{\{\|X\|>T\}}]$, uniformly
in $n$. Letting $T\to\infty$ proves continuity.
\end{proof}

Write $\xi=P-e_Y$. For every other feasible prediction $Q$, the
least-squares projection inequality is
\begin{equation}
 \mathbb E\langle \xi,Q-P\rangle\ge0.
 \label{core:fit-vi}
\end{equation}
It follows by differentiating the squared risk along the feasible
segment $(1-t)P+tQ$. Every nonexpansive map from $\Delta_K$ into itself
can be postcomposed with a fitted profile while preserving its cap.

\begin{lemma}[No residual normal to a simplex face]
\label{core:face}
For every measurable $H$-valued score $S$, every fit $P$ at $S$, and every
$j\in[K]$, we have $\mathbb P(P_j=0,\,Y=j)=0$.
Consequently, for every integrable random vector $v\in H$,
\begin{equation}
 \lim_{t\downarrow0}\mathbb E\left\langle \xi,
 \frac{\Pi_{\Delta_K}(P+tv)-P}{t}\right\rangle
 =\mathbb E\langle \xi,v\rangle.
 \label{core:face-derivative}
\end{equation}
Here $\Pi_{\Delta_K}$ denotes Euclidean projection, and $P+tv$ belongs
to the affine hyperplane containing the simplex for every real $t$.
\end{lemma}
\begin{proof}
For $0<\epsilon<1/K$, let $T_\epsilon$ be Euclidean projection onto
$\{q\in\Delta_K:q_j\ge\epsilon\ \forall j\}$. This is a
nonexpansive map into the simplex, so $Q=T_\epsilon(P)$ is feasible.
Fix a point $p$ with $s$ positive and $z=K-s$ zero coordinates.
For sufficiently small $\epsilon$, the projection sets the zero
coordinates to $\epsilon$ and subtracts $z\epsilon/s$ from each
positive coordinate. Therefore the right derivative is $1$ on the
zero coordinates and $-z/s$ on the positive coordinates.
Since $\sum_j\xi_j=0$ and $\xi_j=-\mathbf1_{\{Y=j\}}$ on $\{P_j=0\}$, its inner product
with $\xi$ is
\[
 -\frac K s\sum_{j:P_j=0}\mathbf1_{\{Y=j\}}\le0.
\]
The feasible point $(1-K\epsilon)P+\epsilon\mathbf1$ shows that
$\|T_\epsilon(P)-P\|/\epsilon$ is uniformly bounded.
Divide \eqref{core:fit-vi} by $\epsilon$ and apply dominated
convergence. The nonpositive displayed random variable must have zero
expectation, hence vanish almost surely. This proves the support claim.

For the derivative identity, fix $p$, a finite $v$, and a residual
$\xi=p-e_y$ whose support is contained in $J(p)$. For sufficiently small
$t>0$, every originally positive coordinate remains positive after
projection. Projection subtracts the same scalar threshold from these
coordinates. Their residuals sum to zero, and the residuals at all other
coordinates vanish. Consequently
\[
 \left\langle \xi,\Pi_{\Delta_K}(p+tv)-p\right\rangle
 =t\langle \xi,v\rangle
\]
for all sufficiently small $t$, depending on $p,v$.
Nonexpansiveness bounds the absolute difference quotient inner product
by $\|\xi\|\|v\|\le\sqrt2\|v\|$. Dominated convergence proves
\eqref{core:face-derivative} for integrable random directions.
\end{proof}

\begin{lemma}[Convex tests and own-score inequality]
\label{core:fit-identities}
Let $S$ be any integrable $H$-valued random vector, so
$\mathbb E\|S\|<\infty$. Every fitted prediction $P$ at $S$ satisfies
\begin{align}
 \mathbb E \xi&=0,\label{core:mean}\\
 \mathbb E\langle \xi,\nabla f(P-u)\rangle&\le0,
       \label{core:smooth-test}\\
 B\mathbb E\langle \xi,S\rangle
 &\ge\mathbb E\langle \xi,P-u\rangle.
       \label{core:own-score}
\end{align}
The second inequality holds for every finite differentiable convex
$f:H\to\mathbb R$ with globally Lipschitz gradient.
\end{lemma}
\begin{proof}
For every fixed $v\in H$, the maps
$p\mapsto\Pi_{\Delta_K}(p\pm tv)$ are feasible nonexpansive
postcompositions. Apply \eqref{core:fit-vi}, divide by $t$, and use
Lemma~\ref{core:face} with both signs. This gives
$\mathbb E\langle \xi,v\rangle=0$. Since $\xi\in H$, it proves
\eqref{core:mean}.

For a convex test, let
\[
 T_t(s)=\operatorname*{arg\,min}_{a\in H}
       \{f(a)+(2t)^{-1}\|a-s\|^2\}.
\]
The minimizer is unique. The first-order condition is
$T_t(s)=s-t\nabla f(T_t(s))$.
Monotonicity of $\nabla f$ implies
$\|T_t(s)-T_t(s')\|^2
\le\langle T_t(s)-T_t(s'),s-s'\rangle$, hence $T_t$ is
nonexpansive. Thus
$Q_t=\Pi_{\Delta_K}(u+T_t(P-u))$ is feasible.
On the compact set $\mathcal C$, the first-order equation and the Lipschitz
gradient bound show, uniformly in $s\in \mathcal C$, that
\[
 \frac{T_t(s)-s}{t}\longrightarrow-\nabla f(s).
\]
For example, if the gradient has Lipschitz constant $L_f$, then for
$tL_f<1/2$ its displacement is bounded by
$2t\sup_{s\in \mathcal C}\|\nabla f(s)\|$; the displayed convergence follows.
Nonexpansiveness of simplex projection allows replacement of these
uniformly converging directions by $-\nabla f(P-u)$.
Dividing \eqref{core:fit-vi} by $t$ and applying
Lemma~\ref{core:face} proves \eqref{core:smooth-test}.

Finally, for $0\le t\le1$ the profile
\[
 s\longmapsto
 \Pi_{\Delta_K}\big(u+(1-t)(\rho(s)-u)+tBs\big)
\]
is $B$-Lipschitz and simplex-valued. Its displacement direction at
$t=0$ is $BS-(P-u)$, which is integrable. The same derivative argument
in \eqref{core:fit-vi} proves \eqref{core:own-score}.
\end{proof}

\subsection{Extending the tests to nonsmooth boundary behavior}

\begin{lemma}[Facewise convex tests]
\label{core:face-test}
Let $P$ and $Y$ be measurable random variables taking values in
$\Delta_K$ and $[K]$, respectively, and put $\xi=P-e_Y$.
Suppose they obey the support conclusion
of Lemma~\ref{core:face} and all the inequalities
\eqref{core:smooth-test}. If $F$ is finite continuous and convex on
$\Delta_K$ and differentiable on the relative interior of every nonempty face, then
\begin{equation}
 \mathbb E\langle \xi,\nabla_{J(P)}F(P)\rangle\le0.
 \label{core:general-test}
\end{equation}
This random inner product is integrable.
\end{lemma}
\begin{proof}
Put $P_\epsilon=(1-\epsilon)P+\epsilon u$.
Extend $s\mapsto F(u+s)$ by $+\infty$ outside $\mathcal C$ and denote this
closed proper convex function by $\overline F$. Its Moreau envelope
\[
 F_\delta(s)=\inf_{a\in H}
 \{\overline F(a)+(2\delta)^{-1}\|a-s\|^2\}
\]
is finite convex and differentiable, with globally Lipschitz gradient.
On every compact subset of the relative interior of $\mathcal C$, its gradients
converge uniformly to those of $\overline F$ as $\delta\downarrow0$.
For completeness, the proximal points converge uniformly to their
arguments on such a compact set, by comparison with $a=s$ and the
bounded range of $F$ on $\mathcal C$. They eventually remain in a compact
interior subset, where the optimality equation identifies the envelope
gradient with $\nabla\overline F$ at the proximal point. The latter
gradient is continuous, and hence uniformly continuous there.

Apply \eqref{core:smooth-test} to
$f(s)=F_\delta((1-\epsilon)s)/(1-\epsilon)$ and take
$\delta\downarrow0$. Uniform convergence on the trimmed simplex gives
\begin{equation}
 \mathbb E\langle \xi,\nabla_HF(P_\epsilon)\rangle\le0.
 \label{core:trimmed-test}
\end{equation}
Fix $p\in\operatorname{relint}\Delta_J$ and $v\in H_J$.
For sufficiently small $h>0$, both $p\pm hv$ belong to $\Delta_J$.
Convexity bounds
$\langle\nabla_HF((1-\epsilon)p+\epsilon u),v\rangle$
between the two difference quotients obtained by perturbing $p$ by
$\pm hv$ and then contracting toward $u$. Their denominators are
$(1-\epsilon)h$. First let $\epsilon\downarrow0$, using continuity
on the closed simplex; then let $h\downarrow0$, using differentiability
on the face. It follows that
\[
 \langle\nabla_HF((1-\epsilon)p+\epsilon u),v\rangle
 \longrightarrow\langle\nabla_JF(p),v\rangle.
\]
The support condition places $\xi$ in $H_{J(P)}$, so this is the needed
pointwise convergence of the products in \eqref{core:trimmed-test}.

Let $\operatorname{osc}(F)=\max_{\Delta_K}F-\min_{\Delta_K}F$. Supporting inequalities
at $e_Y$ and $u$ give, for $0<\epsilon\le1/2$,
\begin{align*}
 \langle \xi,\nabla_HF(P_\epsilon)\rangle
 &=\langle P_\epsilon-e_Y,\nabla_HF(P_\epsilon)\rangle
   +\epsilon\langle P-u,\nabla_HF(P_\epsilon)\rangle\\
 &\ge -\operatorname{osc}(F)-\frac{\epsilon \operatorname{osc}(F)}{1-\epsilon}\ge-2\operatorname{osc}(F).
\end{align*}
Fatou's lemma applied after adding $2\operatorname{osc}(F)$ proves
\eqref{core:general-test}. The same lower bound and the nonpositive
expectation imply integrability.
\end{proof}

\subsection{The loss-independent certificate and fixed point}

Let $\mathcal W$ be a nonempty compact convex set of matrices mapping
$\mathbb R^d$ to $H$. For a fitted prediction at $W\in\mathcal W$, define
\begin{equation}
 \Gamma(W)=\sup_{V\in\mathcal W}
    \mathbb E\langle P_W-e_Y,(W-V)X\rangle.
 \label{core:gap}
\end{equation}
This quantity is finite when $\mathbb E\|X\|<\infty$ and is
nonnegative because $V=W$ is feasible.

\begin{lemma}[Simultaneous matching-loss certificate]
\label{core:certificate}
Use fitting cap $B\ge\beta>0$ and suppose $\mathbb E\|X\|<\infty$.
For every $W,V\in\mathcal W$ and every $\Phi\in\mathcal E_\beta$,
\begin{align}
 &\mathbb E L_\Phi(P_W,Y)-\mathbb E\ell_\Phi(VX,Y)
 \nonumber\\
 &\quad\le\Gamma(W)-
 \mathbb E\big[\Phi(P_W)+G_\Phi(VX)
                       -\langle P_W-u,VX\rangle\big]
 \label{core:certificate-gap}\\
 &\quad\le\Gamma(W)-\frac1{2\beta}
           \mathbb E\|P_W-q_\Phi(VX)\|^2,
 \label{core:certificate-square}
\end{align}
where $q_\Phi(a)=u+\nabla G_\Phi(a)$ for $a\in H$.
The forecast loss is finite almost surely, and both the forecast loss
and the comparator loss are integrable.
\end{lemma}
\begin{proof}
Write $P=P_W$, $\xi=P-e_Y$, and $b=\mathbb E\langle \xi,P-u\rangle$.
Since $B\ge\beta$, the function
$F(p)=\Phi(p)-\|p-u\|^2/(2B)$ is convex.
On $\{J(P)=J\}$, the residual belongs to $H_J$, so
\[
 \langle \xi,\nabla_J\Phi(P)\rangle
 =B^{-1}\langle \xi,P-u\rangle
   +\langle \xi,\nabla_JF(P)\rangle.
\]
Lemmas~\ref{core:boundary} and \ref{core:face-test} imply that the
forecast loss is integrable and
\begin{equation}
 \mathbb E L_\Phi(P,Y)\le-\mathbb E\Phi(P)+b/B.
 \label{core:forecast-bound}
\end{equation}
The comparator loss is integrable because $G_\Phi$ is Lipschitz and
$\mathbb E\|VX\|<\infty$. With
$D_\Phi(P,VX)=\Phi(P)+G_\Phi(VX)-\langle P-u,VX\rangle$,
its risk equals
\[
 -\mathbb E\Phi(P)+\mathbb E D_\Phi(P,VX)
                         +\mathbb E\langle \xi,VX\rangle.
\]
Subtract this identity from \eqref{core:forecast-bound}, use
$b/B\le\mathbb E\langle \xi,WX\rangle$ from
\eqref{core:own-score}, and then use \eqref{core:gap}.
This proves \eqref{core:certificate-gap}.
Finally, the $1/\beta$-strongly convex function
$p\mapsto\Phi(p)-\langle p-u,VX\rangle$ is minimized on the simplex
at $q_\Phi(VX)$. Its objective gap is therefore at least
$\|P-q_\Phi(VX)\|^2/(2\beta)$, proving
\eqref{core:certificate-square}, including at boundary minimizers.
\end{proof}

\paragraph{Proof of Theorem~\ref{thm:main}.}
\begin{proof}
Fit with $B=\beta$. The set $\mathcal W$ is compact and convex in the
finite-dimensional matrix space. By Lemma~\ref{core:fit-continuity},
\[
 W\longmapsto\Pi_{\mathcal W}^{F}\big(W-g(W)\big)
\]
is a continuous self-map of this set, where the superscript indicates
Euclidean projection in Frobenius geometry, even when $\mathcal W$ is defined by another norm or additional convex constraints. Brouwer's theorem supplies a fixed point $W$.
Projection optimality gives
\[
 \langle g(W),V-W\rangle_F\ge0\qquad(V\in\mathcal W).
\]
Since $g(W)=\mathbb E[(P_W-e_Y)X^\top]$, this says exactly that
$\Gamma(W)=0$. Lemma~\ref{core:certificate} proves all the risk
comparisons simultaneously. No part of the fitted prediction or fixed
point construction depends on $\Phi$. The support statement is
Lemma~\ref{core:face}. Finally, put $r=\operatorname{rank}(W)$ and
choose a linear isometry $Q:\R^r\to H$ onto $\operatorname{range}(W)$.
Write $A=Q^\top W$ and $\widetilde\rho=\rho\circ Q$.
Then $W=QA$, $\|A\|_F=\|W\|_F$, and
$\|A\|_{\mathrm{op}}=\|W\|_{\mathrm{op}}$, while
$\widetilde\rho$ retains the Lipschitz cap $\beta$.
Thus the predictor uses $r\le\min\{d,K-1\}$ scalar linear scores.
\end{proof}

\subsection{Finite-action postprocessing}

\begin{corollary}[Finite actions without changing the learned forecast]
\label{revision:finite-actions}
Let $P=\rho(WX)$ be obtained by the least-squares fixed-point
construction in the proof of Theorem~\ref{core:global}, with fitting
cap $\beta$, $W\in\mathcal W$, and $\Gamma(W)=0$.
Fix $\Phi\in\mathcal E_\beta$.
For $0<\delta<1$, define
\[
 p_\delta=(1-\delta)p+\delta u,
 \qquad k_{\Phi,\delta}(p)=\nabla_H\Phi(p_\delta),
\]
and the maximal Jensen gap
\[
 J_\Phi(\delta)=\max_{p\in\Delta_K}
 \big\{(1-\delta)\Phi(p)+\delta\Phi(u)-\Phi(p_\delta)\big\}.
\]
Then $k_{\Phi,\delta}:\Delta_K\to H$ is continuous and bounded.
The postprocessed loss and every comparator loss are integrable,
the comparator infimum is finite, and
\begin{align}
 &\mathbb E\ell_\Phi(k_{\Phi,\delta}(P),Y)\notag\\
 &\quad\le\inf_{V\in\mathcal W}\mathbb E\ell_\Phi(VX,Y)
       +\frac{J_\Phi(\delta)}{1-\delta}
       +\frac{\delta(1-1/K)}{4\beta}.
 \label{revision:smoothing-bound}
\end{align}
Moreover, $J_\Phi(\delta)\to0$ as $\delta\downarrow0$ and is
unchanged by adding an affine function on $\R^K$ to $\Phi$.
Thus, for every $\Phi\in\mathcal E_\beta$ and every $\varepsilon>0$,
there is a $\delta\in(0,1)$ for which the same forecast, with the
finite-action postprocessing $k_{\Phi,\delta}$, has excess risk at
most $\varepsilon$ against the comparator class. No retraining or change to $W$ or $\rho$ is required.
The smoothing parameter may depend on the target entropy.
\end{corollary}
\begin{proof}
Write $\xi=P-e_Y$, $P_\delta=(1-\delta)P+\delta u$, and
$b=\mathbb E\langle \xi,P-u\rangle$. The trimmed convex-test inequality
\eqref{core:trimmed-test}, applied to
$F(p)=\Phi(p)-\|p-u\|^2/(2\beta)$, gives
\[
 \mathbb E\langle \xi,\nabla_H\Phi(P_\delta)\rangle
 \le\frac{1-\delta}{\beta}\,b.
\]
Fenchel equality at the interior probability $P_\delta$ therefore yields
\begin{align*}
 \mathbb E\ell_\Phi(k_{\Phi,\delta}(P),Y)
 &=-\mathbb E\Phi(P_\delta)
      +\mathbb E\langle P_\delta-e_Y,
                         \nabla_H\Phi(P_\delta)\rangle\\
 &\le-\mathbb E\Phi(P_\delta)
       +\frac{1-\delta}{\beta}b
       +\delta\mathbb E\langle u-P,
                         \nabla_H\Phi(P_\delta)\rangle.
\end{align*}
Convexity at $u$ gives
\[
 \delta\langle u-P,\nabla_H\Phi(P_\delta)\rangle
 \le\frac{\delta}{1-\delta}
             \big(\Phi(u)-\Phi(P_\delta)\big).
\]
The fixed-point comparator inequality is
$\mathbb E\ell_\Phi(VX,Y)\ge-\mathbb E\Phi(P)+b/\beta$.
Subtracting it from the preceding bounds gives, for every $V$,
\[
 \mathbb E\ell_\Phi(k_{\Phi,\delta}(P),Y)
 -\mathbb E\ell_\Phi(VX,Y)
 \le\frac{J_\Phi(\delta)}{1-\delta}-\frac{\delta b}{\beta}.
\]

It remains to lower-bound $b$. The fitted mean identity gives
$\mu=\mathbb E P=\mathbb E e_Y$. With
$\eta(X)=\mathbb E[e_Y\mid X]$,
\begin{align*}
 b
 &=\mathbb E\left\|P-\mu-\tfrac12(\eta-\mu)\right\|^2
       -\tfrac14\mathbb E\|\eta-\mu\|^2\\
 &\ge-\tfrac14\mathbb E\|e_Y-\mu\|^2
 =-\tfrac14(1-\|\mu\|^2)
 \ge-\tfrac14(1-1/K).
\end{align*}
This proves \eqref{revision:smoothing-bound} after taking the infimum
over $V$. Continuity of $\Phi$ on the compact simplex gives uniform
convergence $J_\Phi(\delta)\to0$. Affine contributions cancel from its
definition. Finally, all $p_\delta$ lie in the compact trimmed simplex;
the gradient of the differentiable convex function $\Phi$ is continuous
there, and hence bounded. Thus the proposed rule uses finite actions
uniformly over all forecasts and all displayed risks are finite.
\end{proof}

\paragraph{Scalar binary actions.}
For the forecast of Theorem~\ref{thm:binary}, the same calculation gives
the finite action $\varphi'((1-\delta)p+\delta/2)$ with excess at most
$J_\varphi(\delta)/(1-\delta)+\delta/(16\beta)$, where
\[
 J_\varphi(\delta)=\max_{p\in[0,1]}
 \{(1-\delta)\varphi(p)+\delta\varphi(1/2)
       -\varphi((1-\delta)p+\delta/2)\}.
\]
The scalar trimmed convex test and comparator inequality give
$J_\varphi(\delta)/(1-\delta)-\delta b/\beta$, with
$b=\mathbb E[(P-Y)(P-1/2)]$. Pointwise completion of the square gives
$(P-Y)(P-1/2)=(P-1/2-(Y-1/2)/2)^2-1/16\ge-1/16$.
Continuity of $\varphi$ implies $J_\varphi(\delta)\to0$, so finite
actions achieve every positive accuracy without retraining.

\paragraph{A sharper bound for logarithmic loss.}
For $\Phi_T(p)=T\sum_jp_j\log p_j$, the finite rule above has canonical
loss $-T\log((1-\delta)p_y+\delta/K)$. Whenever the original label is
in the support of $p$,
\[
 -T\log((1-\delta)p_y+\delta/K)
 \le-T\log p_y-T\log(1-\delta).
\]
The population support property therefore gives the sharper excess
bound $-T\log(1-\delta)$. In particular,
$\delta=1-\exp(-\varepsilon/T)$ gives excess at most $\varepsilon$
using finite logits. The centered action has coordinate range at most
$T\log(K/\delta)$, so its Euclidean norm is at most
$(T\sqrt K/2)\log(K/\delta)$.

\subsection{The joint score--slope lower bound}

\begin{theorem}[Exact joint score--slope tradeoff]
\label{revision:joint-rank-slope}\label{core:rank-sharp}\label{core:slope-sharp}
Fix integers $K\ge2$ and $d\ge1$, put $d_\star=\min\{d,K-1\}$, and let $R,\beta,L>0$ satisfy
$\beta RL/\sqrt{d_\star}<1/K$. There is one feature--label distribution in $\mathbb R^d$, with
$X$ uniform, with respect to normalized $d_\star$-dimensional Lebesgue measure,
on the open radius-$L$ ball centered at the origin in a $d_\star$-dimensional
linear subspace, and one quadratic
entropy in $\mathcal E_\beta$ with the following property.
For every $r\in\{1,\ldots,d_\star \}$ and $B\ge0$, the smallest regret over
forecasts $p(x)=\rho(Wx)$ with linear $W:\R^d\to H$ and
$\rho:H\to\Delta_K$ satisfying
\[
 \operatorname{rank}(W)\le r,\qquad
 \|W\|_F\le R,\qquad \operatorname{Lip}(\rho)\le B
\]
is exactly
\begin{equation}
 \frac{L^2}{2\beta(d_\star+2)}
 \left[(d_\star-r)c^2+r\left(c-\frac{BR}{\sqrt r}\right)_+^2\right],
 \qquad c=\frac{\beta R}{\sqrt{d_\star}}.
 \label{revision:joint-value}
\end{equation}
The benchmark is the finite infimum of expected matching loss over
linear $V:\R^d\to H$ with $\|V\|_F\le R$; the head's Lipschitz
constant is measured in Euclidean norms, and $(a)_+=\max\{a,0\}$.
For $r=0$ and every $B\ge0$, the value is
$L^2d_\star c^2/(2\beta(d_\star+2))$. All these values are attained by feasible forecasts that are affine
on the closed radius-$L$ ball in the feature-support subspace and
have globally Lipschitz simplex-valued heads.
\end{theorem}
Here a rank-at-most-$r$ map into $H$ is equivalent to a forecast based
on $r$ scalar scores. Indeed, choose an isometry
$Q_0:\mathbb R^r\to H$ whose range contains $\operatorname{range}(W)$.
Then $W=Q_0A$ with $\|A\|_F=\|W\|_F$, and the head
$t\mapsto\rho(Q_0t)$ has the same Lipschitz cap. Conversely, an
$r$-score head $h:\mathbb R^r\to\Delta_K$ is represented on $H$ by
$s\mapsto h(Q_0^\top s)$, again without increasing the cap.
\begin{proof}
Identify the support subspace with $\mathbb R^{d_\star}$. Restrict each candidate score matrix to this subspace; its norm and rank cannot increase. Conversely, extend each attaining matrix by zero on the orthogonal complement. We henceforth work in these $d_\star$ coordinates and fix an isometry $U:\mathbb R^{d_\star}\to H$. Take $X$ uniform on the
radius-$L$ ball, let $\eta(x)=u+cUx$ on this ball, and conditionally draw
$Y\sim\eta(X)$. The radius condition makes every class probability
strictly positive on the closed support ball. Use $\Phi(p)=\|p-u\|^2/(2\beta)$.
The comparator $V_0=(R/\sqrt{d_\star})U$ has Frobenius norm $R$ and is a
conditional Bayes action, since $u+\nabla G_\Phi(V_0x)=\eta(x)$.
The comparator benchmark is therefore $-\mathbb E\Phi(\eta(X))$.
For an interior forecast, its conditional excess risk is
$\|p(x)-\eta(x)\|^2/(2\beta)$. Boundary forecasts have infinite
conditional extended loss, so the same squared-error expression
remains a lower bound for arbitrary forecasts.

Consider any feasible $W$ and choose an orthogonal matrix $Q$ whose
columns are its right singular vectors. Write
$\sigma_j=\|WQe_j\|$, arranging the possibly nonzero singular values
among $j\le r$; then $\sigma_j=0$ for $j>r$ and
$\sum_{j\le r}\sigma_j^2\le R^2$. In the rotated coordinates
$T=Q^\top X$, define
\[
 f_j(t)=\langle UQe_j,p(Qt)-u\rangle,
 \qquad v=\mathbb E T_j^2=\frac{L^2}{d_\star+2}.
\]
Let $Q_j$ reflect the $j$th coordinate of $\mathbb R^{d_\star}$. Since
$Q_jT$ has the same law as $T$, and the scalar output coordinate has
Lipschitz constant at most $B\sigma_j$ along this direction,
\begin{align*}
 \mathbb E[T_jf_j(T)]
 &=\frac12\mathbb E\big[T_j(f_j(T)-f_j(Q_jT))\big]\\
 &\le B\sigma_j\,\mathbb E T_j^2=B\sigma_j v.
\end{align*}
Set $a_j=\mathbb E[T_jf_j(T)]/v$. Cauchy--Schwarz gives
$\mathbb E f_j(T)^2\ge v a_j^2$. Hence
\[
 \mathbb E(f_j(T)-cT_j)^2
 \ge v(a_j-c)^2\ge v(c-B\sigma_j)_+^2.
\]
The last inequality uses $a_j\le B\sigma_j$; when $B\sigma_j\ge c$
its right-hand side is zero. The vectors $UQe_j$ are orthonormal in $H$. Summing their squared errors is therefore a lower bound on the total squared error, and gives the regret lower bound
\[
 \frac v{2\beta}\left[(d_\star-r)c^2+
                  \sum_{j=1}^r(c-B\sigma_j)_+^2\right].
\]
The function $h(t)=(c-B\sqrt t)_+^2$ is convex and nonincreasing on
$[0,\infty)$: before its zero threshold its derivative increases from
$-\infty$ to zero, and thereafter it is constant (the case $B=0$ is
constant throughout). Jensen's inequality and the Frobenius constraint
therefore imply
\[
 \sum_{j=1}^r h(\sigma_j^2)
 \ge r h\left(\frac1r\sum_{j=1}^r\sigma_j^2\right)
 \ge r h(R^2/r).
\]
This is the lower bound in \eqref{revision:joint-value}.

For attainment, fix an $r$-dimensional subspace $E\subseteq\mathbb R^{d_\star}$
and put $a=\min\{c,BR/\sqrt r\}$. Define
\[
 W=\frac R{\sqrt r}U\Pi_E,
 \qquad
 \rho(s)=\Pi_{\Delta_K}
    \left(u+\frac{a\sqrt r}{R}\Pi_{UE}s\right).
\]
Then $\operatorname{rank}(W)=r$, $\|W\|_F=R$, and
$\operatorname{Lip}(\rho)\le a\sqrt r/R\le B$.
For $\|x\|\le L$, the probability vector $u+aU\Pi_Ex$ is strictly
interior because $aL\le cL<1/K$. Thus projection does not alter it,
and the realized forecast is $p(x)=u+aU\Pi_Ex$. Isotropy gives
\[
 \mathbb E\|p(X)-\eta(X)\|^2
 =v\big[(d_\star-r)c^2+r(c-a)^2\big],
\]
which attains \eqref{revision:joint-value}. For $r=0$, use $p\equiv u$.
\end{proof}

\paragraph{Arbitrary measurable heads.}
The rank lower bound does not require a finite Lipschitz cap.
Continue in the $d_\star$-dimensional support coordinates, and let $W$
denote the restriction of the original score map to that subspace.
For any measurable $p(x)=\rho(Wx)$, put
$E=\operatorname{range}(W^\top)\subseteq\R^{d_\star}$.
Reflection symmetry of the ball gives $\mathbb E[X\mid WX]=\Pi_E X$.
Conditional least squares therefore implies
\[
 \mathbb E\|p(X)-\eta(X)\|^2
 \ge c^2\mathbb E\|X-\Pi_E X\|^2
 =c^2(d_\star-\dim E)\frac{L^2}{d_\star+2}.
\]
The interior forecast $u+cU\Pi_E X$ attains this bound. The quadratic
finite canonical action has the same squared-error identity even at
boundary reports, so the joint tradeoff also holds under that convention.

\paragraph{Consequences.}
At $B=\beta$, \eqref{revision:joint-value} reduces to \eqref{eq:rank-regret}. At full score rank $r=d_\star$, it becomes
\[
 \frac{R^2L^2}{2\beta(d_\star+2)}(\beta-B)_+^2.
\]
Thus one distribution quantitatively witnesses both necessary score
dimension and necessary learned slope. More generally, when only $r$
directions are retained, no additional slope penalty is paid once
$B\ge\beta\sqrt{r/d_\star}$: the available Frobenius budget can be
concentrated on those retained directions.

\section{Separate scalar probability links cannot replace a joint multiclass link}
\label{separate:section}

The multiclass representation uses $K-1$ linear scores jointly:
$p(x)=\rho(Wx)$, where $\rho:\mathbb R^{K-1}\to\Delta_K$.
It is natural to ask whether this can be replaced by $K-1$ scalar
nondecreasing links,
\begin{equation}
 p_j(x)=\rho_j(a_j^\top x)\quad(j\le K-1),
 \qquad p_K(x)=1-\sum_{j<K}p_j(x).
 \label{separate:architecture}
\end{equation}
Here $x,a_j\in\R^d$ and $\rho_j:\R\to[0,1]$; the right side must be
a probability vector almost surely under the law of $X$.
The links are measurable (as every nondecreasing real function is).
If necessary, replace the forecast by $u$ on the measurable null set
where simplex membership fails. This produces an everywhere
simplex-valued measurable forecast with the same almost-sure identities.
The answer is negative even if the scalar links are allowed to be
arbitrary measurable functions, with no slope or coefficient restriction.
Indeed, requiring even one class probability to be a function of a
single linear score can force positive regret.

\begin{theorem}[A joint probability link is necessary]
\label{separate:impossibility}
Fix an integer $K\ge3$ and $R,\beta,L>0$.
There is a joint distribution on $\R^{K-1}\times[K]$ with $X$ uniform
on the open radius-$L$ Euclidean ball centered at the origin, and
one entropy $\Phi\in\mathcal E_\beta$, with the following properties.
The potential $G_\Phi:H\to\R$ is differentiable, and
$\ell_\Phi(VX,Y)$ is integrable for every linear $V:\R^{K-1}\to H$.
For a linear $V_0:\R^{K-1}\to H$, its conditional probability vector is
\[
 \eta(x)=u+\nabla G_\Phi(V_0x),\qquad \|V_0\|_F=R,
\]
where $\nabla G_\Phi$ is the gradient on $H$.
The radius-$R$ linear benchmark is Bayes-optimal and finite:
\[
 \E L_\Phi(\eta(X),Y)
 =\inf_{\|V\|_F\le R}\E\ell_\Phi(VX,Y).
\]
Moreover, $\eta=\rho\circ W$ for a linear
$W:\R^{K-1}\to\R^{K-1}$ with $\|W\|_F\le R$ and a joint
$\beta$-Lipschitz map $\rho:\R^{K-1}\to\Delta_K$.
There is a constant $\gamma>0$, depending only on $K,R,\beta,L$,
such that every measurable forecast $p:\mathbb R^{K-1}\to\Delta_K$
with
\[
 p_j(x)=f(a^\top x)\quad\text{almost surely}
\]
for at least one class $j$, some $a\in\mathbb R^{K-1}$, and some
measurable $f:\mathbb R\to[0,1]$, satisfies
\begin{equation}
 \mathbb E L_\Phi(p(X),Y)
 \ge \inf_{\|V\|_F\le R}\mathbb E\ell_\Phi(VX,Y)+\gamma.
 \label{separate:gap}
\end{equation}
The forecast risk in this inequality is an ordinary extended
expectation and may be $+\infty$.
Consequently, the exact existence theorem cannot in general retain
the separate-link architecture \eqref{separate:architecture}, even
without requiring its scalar links to be nondecreasing or Lipschitz.
\end{theorem}

\begin{proof}
Put $m=K-1$, choose an isometry $U:\mathbb R^m\to H$, and set
\[
 T=\frac1{2\beta},\qquad
 V_0=\frac R{\sqrt m}U,\qquad
 \Phi(q)=T\sum_{j=1}^Kq_j\log q_j.
\]
As shown for the Shannon examples, $\Phi\in\mathcal E_\beta$ and
its probability link is $q_\Phi(a)=\operatorname{softmax}(a/T)$,
with Euclidean Lipschitz constant at most $1/(2T)=\beta$.
Let $X$ have the stated uniform law and, conditionally on $X=x$,
draw $Y$ with probabilities
\[
 \eta(x)=\operatorname{softmax}(\alpha Ux),
 \qquad \alpha=\frac{R}{T\sqrt m}>0.
\]
All class probabilities are strictly positive. The admissible comparator
$V_0$ is a conditional Bayes action, and $\|V_0\|_F=R$.
For every probability forecast, with infinite values permitted,
\begin{equation}
 \mathbb E L_\Phi(p(X),Y)
   -\inf_{\|V\|_F\le R}\mathbb E\ell_\Phi(VX,Y)
 =T\mathbb E\operatorname{KL}(\eta(X)\|p(X)).
 \label{separate:kl}
\end{equation}

We first show that no coordinate $\eta_j$ is a scalar ridge function
on the feature ball. Write $h_j=U^\top e_j$, so
$\|h_j\|^2=(K-1)/K$. Direct differentiation at zero gives
\begin{equation}
 \nabla^2\eta_j(0)
 =\frac{\alpha^2}{K}
       \left(h_jh_j^\top-\frac1K I_m\right).
 \label{separate:hessian}
\end{equation}
Its eigenvalue along $h_j$ is $\alpha^2(K-2)/K^2$, and its
remaining $m-1$ eigenvalues are $-\alpha^2/K^2$. It is therefore
nonsingular when $K\ge3$.

In contrast, a smooth function that equals a measurable function of
one linear projection almost everywhere on an open ball has Hessian
rank at most one there. To justify this statement without regularity
of the scalar function, use coordinates $(t,z)$ along the projection
and its orthogonal complement. Fubini's theorem says that, for almost
every $t$, the smooth function is constant almost everywhere on the
corresponding $z$-section. Continuity makes it constant on that whole
section. Continuity of its first derivatives then makes every
orthogonal directional derivative vanish on the entire open ball.
Its Hessian consequently annihilates the orthogonal complement of
the projection direction. A zero projection would make the function
constant. Both possibilities contradict \eqref{separate:hessian}.

To obtain a uniform positive gap, for a unit vector $v$ define
\[
 d_j(v)=\inf_{g}\left(\mathbb E
       |\eta_j(X)-g(v^\top X)|^2\right)^{1/2},
\]
where the infimum is over measurable square-integrable functions of
the displayed score. This is distance to a closed subspace of
$L^2$, and its square is
$\mathbb E\operatorname{Var}(\eta_j(X)\mid v^\top X)$.
The preceding argument proves $d_j(v)>0$ for every $v$.
This distance is continuous in $v$. Indeed, if $v_n\to v$, choose
orthogonal maps $Q_n\to I$ with $Q_nv=v_n$. Rotational invariance
of $X$ gives
\[
 |d_j(v_n)-d_j(v)|
 \le\|\eta_j(Q_nX)-\eta_j(X)\|_{L^2}\longrightarrow0,
\]
by uniform continuity of $\eta_j$ on the closed ball.
Compactness of the unit sphere, and the finite number of classes,
therefore imply
\[
 c_*:=\min_{j\in[K]}\min_{\|v\|=1}d_j(v)^2>0.
\]
Zero-score constant forecasts obey the same lower bound because
constants belong to every one-score subspace.

Grouping class $j$ against all other classes in the log-sum
inequality, followed by the binary Pinsker inequality, gives
\[
 \operatorname{KL}(\eta(x)\|p(x))
 \ge2|\eta_j(x)-p_j(x)|^2.
\]
If $p_j$ is a scalar ridge function, its expected squared error is
at least $c_*$. Substitution in \eqref{separate:kl} proves
\eqref{separate:gap} with $\gamma=2Tc_*>0$.
\end{proof}

\paragraph{What the obstruction does and does not say.}
The restriction in \eqref{separate:architecture} applies to class
\emph{probabilities}. It is different from applying scalar links to
logits and subsequently normalizing them. For example,
\[
 p_j(x)=\frac{\exp(\rho_j(a_j^\top x))}
 {1+\sum_{\ell<K}\exp(\rho_\ell(a_\ell^\top x))}\quad(j<K),
 \qquad
 p_K(x)=\frac1{1+\sum_{\ell<K}\exp(\rho_\ell(a_\ell^\top x))}
\]
has a joint probability link even when all $\rho_j$ are scalar and
nondecreasing. Such an architecture includes linear softmax and is
not excluded by Theorem~\ref{separate:impossibility}.
Likewise, this theorem does not settle whether the learned joint
link can satisfy a specified multivariate monotonicity condition,
such as $\langle\rho(s)-\rho(t),s-t\rangle\ge0$ on $H$.
The general existence proof does not establish that additional condition.
The Dirichlet construction in Section~\ref{sec:dirichlet-learning} does
satisfy it under its additional feature assumption.
In score coordinates $\mathbb R^{K-1}$, requiring all $K$ probability
coordinates to be nondecreasing in every input coordinate would force a
constant map: on
comparable inputs all coordinate increments are nonnegative and
sum to zero, and any two inputs have a common coordinatewise upper
bound. Requiring monotonicity only for the first $K-1$ outputs is
a different restriction and is not addressed by this argument.

\section{Statistical learning with a fixed number of scores}
\label{stat:section}\label{sec:statistics-appendix}

This appendix gives a dimension-free statistical consequence of the
representation theorem. It requires additional regularity of the target
entropies. In particular, Theorem~\ref{thm:pac} does not cover the logarithmic
loss at the boundary of the simplex.

Fix $K\ge2$ and $L,R,\beta,\lambda,M>0$, and put $m=K-1$,
$u=\mathbf 1/K$, and
$H=\{a\in\mathbb R^K:\langle a,\mathbf 1\rangle=0\}$.
Let $\mathcal X$ be a separable real Hilbert space with its Borel
$\sigma$-algebra, and let $(X,Y)$ have a Borel probability law on
$\mathcal X\times[K]$ with $\|X\|\le L$ almost surely. Set $A=LR$.
All operators $W:\mathcal X\to H$ are bounded and linear, and
$\|W\|_F$ denotes their Hilbert--Schmidt norm,
$\|W\|_F^2=\sup_E\sum_{e\in E}\|We\|^2$, where $E$ ranges over
finite orthonormal subsets of $\mathcal X$.
Write $\mathcal E(\beta,\lambda,M)$ for
the smooth family defined above. In particular, its relative gradients satisfy
\begin{equation}
 \sup_{q\in\Delta_K}\|\nabla_H\Phi(q)\|\le M,
 \qquad
 \|\nabla_H\Phi(q)-\nabla_H\Phi(q')\|
 \le\lambda\|q-q'\|.
 \label{stat:regularity}
\end{equation}
The matching potential and the finite canonical loss are
\begin{align}
 G_\Phi(a)&=\max_{q\in\Delta_K}
       \{\langle q-u,a\rangle-\Phi(q)\},\nonumber\\
 \ell_\Phi(a,y)&=G_\Phi(a)-\langle e_y-u,a\rangle,\nonumber\\
 \mathcal L_\Phi(q,y)
 &=\ell_\Phi(\nabla_H\Phi(q),y)\nonumber\\
 &=\langle q-e_y,\nabla_H\Phi(q)\rangle-\Phi(q).
 \label{stat:canonical}
\end{align}
Fenchel equality justifies the last identity, including at simplex
faces. This is a finite canonical selector, rather than the extended
face loss used for possibly singular entropies in the existence theorem.

We prove Theorem~\ref{thm:pac}; the proof also allows any fitting
cap $B\ge\beta$. In that extension set $D=M+A+B\lambda A$.
For $0<\varepsilon\le D$, $0<\delta<1/2$, and integer $n\ge1$,
the sufficient sample bound is
\begin{equation}
 n\ge
 \begin{cases}
 C_K(D/\varepsilon)^2\log^2\!\big(C_KD/(\varepsilon\delta)\big),&m\le2,\\
 C_K\big[(D/\varepsilon)^m+(D/\varepsilon)^2\log(2/\delta)\big],&m\ge3.
 \end{cases}
 \label{stat:sample-bound}
\end{equation}
The desired simultaneous guarantee is
\begin{equation}
 \E\mathcal L_\Phi(p(X),Y)
 \le\inf_{\|V\|_F\le R}\E\ell_\Phi(VX,Y)+\varepsilon.
 \label{stat:guarantee}
\end{equation}
The learner is a finite search; its general running time need not be
polynomial. The higher-class accuracy exponent is not claimed optimal.

\subsection{An empirical certificate}
\label{stat:empirical-section}
Write $\mathcal C=\Delta_K-u$, $Z=e_Y-u$, and let $\widehat{\mathbb E}$
denote averaging over a sample. For a score matrix $W$, let
$\widetilde P_W=\widetilde\rho_W(WX)\in\mathcal C$ be a least-squares fit to $Z$ among
all $B$-Lipschitz maps from the radius-$A$ ball of $H$ into
$\mathcal C$. Write $P_W=u+\widetilde P_W$ for the corresponding probability fit. The fitted sample vector is unique because the feasible
sample values form a compact convex set. Define
\begin{align}
 g_n(W)&=\widehat{\mathbb E}[(\widetilde P_W-Z)X^*],\nonumber\\
 \Gamma_n(W)&=\langle g_n(W),W\rangle_F+R\|g_n(W)\|_F.
 \label{stat:gap}
\end{align}
Here $X^*$ is the functional $v\mapsto\langle X,v\rangle$.

\begin{lemma}[Empirical simultaneous comparison]
\label{stat:empirical}
For every sample of size $n\ge1$ with $\|X_i\|\le L$, the unique fitted sample vector at each $W$ with $\|W\|_F\le R$ satisfies
\begin{equation}
 \widehat{\mathbb E}\mathcal L_\Phi(P_W,Y)
 \le\inf_{\|V\|_F\le R}
      \widehat{\mathbb E}\ell_\Phi(VX,Y)+\Gamma_n(W)
 \label{stat:empirical-bound}
\end{equation}
for all $\Phi\in\mathcal E(\beta,\lambda,M)$ simultaneously.
There exists such a $W$ with $\Gamma_n(W)=0$ that vanishes on $(\operatorname{span}\{X_1,\ldots,X_n\})^\perp$.
\end{lemma}
\begin{proof}
Apply Lemma~\ref{core:certificate} to the empirical distribution,
with fitting cap $B$. The finite pairwise Lipschitz constraints extend
to a global link by Lemma~\ref{stat:same-cap-extension}, so their
least-squares fit is the same fit used by that certificate. Its support
property implies that the finite canonical loss agrees with the
facewise loss on every observed label. The gap in~\eqref{core:gap}
over the Frobenius ball equals~\eqref{stat:gap}. This proves the
empirical inequality.

Restrict the score matrices to the finite-dimensional sample span.
Orthogonal projection onto this span preserves every observed score and
cannot increase the Frobenius norm. Lemma~\ref{core:fit-continuity}
gives continuity of the fitted sample vector and of $g_n$.
Brouwer's theorem applied to $W\mapsto\Pi_{\|V\|_F\le R}(W-g_n(W))$
therefore gives a sample-span matrix with $\Gamma_n(W)=0$.
For a separable Hilbert feature space, this argument remains entirely
inside the same finite-dimensional sample span.
\end{proof}

\subsection{Uniform convergence for Lipschitz profiles}
\label{stat:uniform-section}
For $C,M_0>0$, define $\mathcal F_{m,C,M_0}$ to contain functions on $\{x\in\mathcal X:\|x\|\le L\}\times[K]$ of the form
\[
 f(x,y)=a_y(Wx),\qquad \|W\|_F\le R,
\]
where each $a_y$ is $C$-Lipschitz on the radius-$A$ score ball and
takes values in $[-M_0,M_0]$. Extend $f$ by zero for $\|x\|>L$.
These profiles are not required to arise from a loss or a probability forecast.

\begin{lemma}[Dimension-free profile bound]
\label{stat:uniform}
Let $n\ge1$ and $0<\delta<1$, and set $D_0=M_0+CA$. For an i.i.d. sample from the preceding law, on a measurable event of probability at least $1-\delta$,
\[
 \sup_{f\in\mathcal F_{m,C,M_0}}
   |\mathbb Ef-\widehat{\mathbb E}f|
 \le
 \begin{cases}
 \displaystyle
 C_K\frac{D_0\log(2n)+M_0\sqrt{\log(2/\delta)}}{\sqrt n},&m\le2,\\[3mm]
 \displaystyle
 C_KD_0n^{-1/m}
       +C_KM_0\sqrt{\frac{\log(2/\delta)}n},&m\ge3.
 \end{cases}
 \label{stat:uniform-bound}
\]
The bound is independent of the dimension of $\mathcal X$.
\end{lemma}
\begin{proof}
The supremum is measurable. Since the feature Hilbert space is separable and $H$ is finite dimensional, the Hilbert--Schmidt radius-$R$ ball has a countable dense subset. Bounded Lipschitz profiles on the compact score ball likewise admit a countable uniform-dense subset. Their compositions are uniformly dense on the bounded feature domain: an operator error $h$ changes an original profile by at most $CLh$. The displayed supremum therefore equals a countable supremum of measurable deviations.
We first bound empirical covering numbers in normalized Euclidean
sample distance. The image of the Frobenius radius-$R$ ball under
$W\mapsto(Wx_i)_{i=1}^n$ is an ellipsoid. In this distance its squared
semiaxes sum to at most $mR^2n^{-1}\sum_i\|x_i\|^2\le mA^2$.
An ellipsoid with squared semiaxis sum $S$ has logarithmic covering
number at most $cS/t^2$ at scale $t$. Here is a direct internal-cover
argument. Retain the axes $a_j>t/4$ and call the resulting ellipsoid
$E_N$; the discarded component has norm at most $t/4$. A maximal
$t/2$-separated set in $E_N$ is an internal $t/2$-cover, and the
radius-$t/4$ balls around its centers are disjoint. For $s=t/4$,
\[
 E_N+sB_2\subseteq\sqrt2\,\operatorname{diag}(a_j+s)B_2.
\]
Indeed, for $\|v\|,\|w\|\le1$, coordinatewise convexity of the square
bounds $\sum_j(a_jv_j+sw_j)^2/(a_j+s)^2$ by
$\sum_jv_j^2+\sum_jw_j^2\le2$.
Comparing volumes therefore bounds the number of separated points by
$\prod_j\sqrt2(1+4a_j/t)$. Because $a_j/t>1/4$, its logarithm is at
most $c\sum_j a_j^2/t^2$. Extending each center by zero on the omitted
axes keeps it inside the original ellipsoid and gives a $t$-cover.
Thus the score image has log-cover size at most $c mA^2/t^2$.

A scalar $C$-Lipschitz function on the score ball admits a uniform
$t$-cover with logarithmic size at most
\[
 c_m\big\{1+(CA/t)^m+\log(1+M_0/t)\big\}.
\]
One direct construction extends it, with the same Lipschitz constant,
to the containing cube and clips the extension to $[-M_0,M_0]$.
Use a regular mesh of spacing $c_m t/C$ and round function values to
multiples of $c_m t$. The first rounded value has
$O(1+M_0/t)$ choices. Along a spanning tree of neighboring mesh vertices,
every subsequent value has only a dimension-dependent constant number
of choices, by Lipschitz continuity. Piecewise constant interpolation
on mesh cells approximates the original function uniformly within $t$.
For scales larger than $CA$, a constant approximation suffices.

Combining $K$ such profile covers at scale proportional to $t$ with a
score cover at scale proportional to $t/C$ yields
\begin{equation}
 \log N(\mathcal F_{m,C,M_0},\|\cdot\|_n,t)
 \le c_K\big\{(CA/t)^2+(CA/t)^m
                    +1+\log(1+M_0/t)\big\}.
 \label{stat:cover}
\end{equation}
When replacing a score by a net center, Lipschitz continuity of the
original profile controls the change before the uniform profile
approximation is applied. Thus the covering profiles themselves need
not retain Lipschitz continuity.

Here are the resulting chaining details. Symmetrization bounds the
expected uniform deviation by twice the expected Rademacher supremum.
For a bounded sample class, approximating at scales $2^{-j}M_0$ and
bounding each finite family of Rademacher increments by its maximum
norm times the square root of twice its logarithmic size gives
\[
 \mathfrak R_n(\mathcal F_{m,C,M_0})
 \le 4\alpha+\frac{c}{\sqrt n}
  \int_\alpha^{2M_0}
     \sqrt{\log N(\mathcal F_{m,C,M_0},\|\cdot\|_n,t)}\,dt
       +\frac{cM_0}{\sqrt n}.
\]
This follows equivalently by summing the dyadic increment bounds; the
unresolved fine-scale remainder has empirical norm at most $2\alpha$.
For $m\le2$, insert \eqref{stat:cover} and stop at
$\alpha=\min\{M_0,D_0/\sqrt n\}$ to obtain
$c_KD_0\log(2n)/\sqrt n$; if $\alpha=M_0$, use the trivial boundedness
estimate. For $m\ge3$, the cover is bounded by
$c_K(D_0/t)^m$ for $0<t\le2M_0$, up to changing the constant.
Stopping at $\alpha=\min\{M_0,D_0n^{-1/m}\}$ gives
$c_KD_0n^{-1/m}$ by integrating $t^{-m/2}$.
Finally, changing one observation changes the uniform deviation by at
most $2M_0/n$. McDiarmid's bounded-difference inequality
\citep{mcdiarmid1989} contributes
$cM_0\sqrt{\log(2/\delta)/n}$ and proves the claim.
\end{proof}

\subsection{A finite quadratic-program extension with the same cap}
\label{stat:same-cap-section}

The extension step can retain the fitted Lipschitz constant exactly.
The following finite-data formula is a quadratic-program form of the
explicit Kirszbraun extension of \citet{azagra2021}, refining the classical extension theorem of \citet{kirszbraun1934}; we give an elementary
proof tailored to probability forecasts. Thus the extension theorem
itself is not a novelty claim of this paper.

\begin{lemma}[A simplex-valued extension without slope inflation]
\label{stat:same-cap-extension}
Let $m\ge0$ and $n\ge1$ be integers, and let $s_1,\ldots,s_n\in\mathbb R^m$ and
$p_1,\ldots,p_n\in\Delta_K$ satisfy
\[
 \|p_i-p_j\|\le B\|s_i-s_j\|\qquad(i,j\in[n]),
\]
where $B>0$. Put $S=[s_1\ \cdots\ s_n]$ and
$P=[p_1\ \cdots\ p_n]$, and define
$c_i=B^2\|s_i\|^2-\|p_i\|^2$. For every $s\in\mathbb R^m$, choose
\begin{equation}
 \theta_s\in\arg\min_{\theta\in\Delta_n} Q_s(\theta),
 \qquad
 Q_s(\theta)=B^2\|S\theta-2s\|^2+\|P\theta\|^2+c^\top\theta,
 \label{stat:extension-qp}
\end{equation}
and let $\rho(s)=P\theta_s$. Then $\rho$ is well defined independently
of the choice of minimizer, takes values in
$\operatorname{conv}\{p_1,\ldots,p_n\}\subseteq\Delta_K$, satisfies
$\rho(s_i)=p_i$, and is $B$-Lipschitz on all of $\mathbb R^m$.
If $\zeta\ge0$ and $\widehat\theta\in\Delta_n$ satisfy
$Q_s(\widehat\theta)\le\min_{\theta\in\Delta_n}Q_s(\theta)+\zeta$, then
\begin{equation}
 \|P\widehat\theta-\rho(s)\|\le\sqrt\zeta.
 \label{stat:extension-accuracy}
\end{equation}
\end{lemma}
\begin{proof}
The objective is continuous and convex on the compact simplex, so it
has a minimizer. If $\theta$ and $\theta'$ are both optimal at the same
query, their midpoint is feasible and the quadratic identity gives
\[
 Q_s\left(\frac{\theta+\theta'}2\right)
 =\frac{Q_s(\theta)+Q_s(\theta')}2
  -\frac14\big(B^2\|S(\theta-\theta')\|^2
                 +\|P(\theta-\theta')\|^2\big).
\]
Optimality forces both squared terms to vanish. In particular the
reported probability is unique, and it belongs to the stated convex hull.

At $s=s_j$, direct differentiation at the simplex vertex $e_j$ gives
\[
 [\nabla Q_{s_j}(e_j)]_i-[\nabla Q_{s_j}(e_j)]_j
 =B^2\|s_i-s_j\|^2-\|p_i-p_j\|^2\ge0.
\]
This is the first-order optimality condition for $e_j$ over $\Delta_n$.
Thus $e_j$ is a minimizer and $\rho(s_j)=p_j$.

For two queries $s,t$, let $\delta=\theta_s-\theta_t$. Add the two
first-order optimality inequalities, testing each minimizer against the
other. Since the coefficient vector $c$ is the same at both queries, it cancels,
leaving
\[
 B^2\|S\delta\|^2+\|P\delta\|^2
 \le 2B^2\langle s-t,S\delta\rangle.
\]
Equivalently,
\[
 \|P\delta\|^2+B^2\|S\delta-(s-t)\|^2
 \le B^2\|s-t\|^2.
\]
Dropping the second nonnegative term proves the Lipschitz claim.

Finally, expand the quadratic objective around a minimizer $\theta_s$.
First-order optimality implies
\[
 Q_s(\widehat\theta)-Q_s(\theta_s)
 \ge B^2\|S(\widehat\theta-\theta_s)\|^2
       +\|P(\widehat\theta-\theta_s)\|^2,
\]
which proves \eqref{stat:extension-accuracy}.
\end{proof}

\paragraph{Finite evaluation and its precision.}
The exact learned head is the unique probability map specified by
\eqref{stat:extension-qp}. It is represented by the finite arrays $S,P,c$.
To evaluate it numerically, solve this convex quadratic program over a
simplex. If $\|s_i\|\le A$, the objective gradient has Lipschitz constant
at most $2n(B^2A^2+1)$ as a function of $\theta$. Projected gradient
descent, starting at any simplex point and using this smoothness bound,
reaches objective error $\zeta$ in
$O(n(B^2A^2+1)/\zeta)$ iterations, since the simplex has squared diameter
at most $2$. Each iteration consists of matrix--vector products and
Euclidean projection onto the simplex, all polynomial-time operations.
The prediction error is at most $\sqrt\zeta$ by
\eqref{stat:extension-accuracy}. Under the statistical entropy assumptions,
every finite canonical loss is $\sqrt2\lambda$-Lipschitz in probability,
so the additional loss error is at most $\sqrt2\lambda\sqrt\zeta$,
uniformly over the target family. The exact optimization-defined map
has the stated cap; arbitrary approximate pointwise evaluations are
only asserted to approximate that map, not to define a globally
$B$-Lipschitz function themselves.

\subsection{A finite learner and transfer to the population}
\label{stat:finite-section}
For fixed sample scores $s_i=Wx_i$, the fitted values are obtained by
\begin{equation}
 \min_{\widetilde p_i\in\mathcal C}\frac1{2n}\sum_i\|\widetilde p_i-(e_{y_i}-u)\|^2,
 \qquad
 \|\widetilde p_i-\widetilde p_j\|\le B\|s_i-s_j\|\quad(i,j\in[n]).
 \label{stat:socp}
\end{equation}
This finite convex program has a unique solution. Its pairwise constraints
are exactly the restrictions of $B$-Lipschitz simplex-valued maps:
Lemma~\ref{stat:same-cap-extension} extends the fitted probabilities
$p_i=u+\widetilde p_i$ with the same constant. Use that lemma's quadratic program
\eqref{stat:extension-qp} as the finite prediction rule. It interpolates
every training value, so all empirical risks and the gap in
\eqref{stat:gap} are preserved, including the vector Lipschitz constant.

For completeness, an approximate empirical fixed point can be found
by a finite search. If $\|W-V\|_F\le h$, transfer the fitted profile at
one matrix to the other and add the two least-squares projection
inequalities. The transferred predictions differ by at most $BLh$,
and residual norms are at most $\sqrt2$, giving
\[
 \|\widetilde P_W-\widetilde P_V\|_{L^2(P_n)}^2\le2\sqrt2 BLh.
\]
Consequently
\[
 |\Gamma_n(W)-\Gamma_n(V)|
 \le\sqrt2 Lh+2A\sqrt{2\sqrt2 BLh}.
\]
A net of the sample-span matrix ball with mesh
\[
 h=c\min\{R,\eta/L,\eta^2/(BA^2L)\}
\]
therefore contains a matrix of gap at most $\eta$, by
Lemma~\ref{stat:empirical}. If the sample span has dimension $r_n\le n$,
choose a finite rectangular grid on the cube containing the matrix ball, fine
enough that its Euclidean covering radius is at most $h/2$, and project
its points onto the ball. In fixed order retain each point at distance
greater than $h/2$ from all previously retained points. The retained points
form an $h$-net. Their disjoint open balls of radius $h/4$ lie inside the
radius-$(R+h/4)$ ball, so the net has at most
\[
 \big(C\max\{1,A/\eta,BA^3/\eta^2\}\big)^{mr_n}
\]
points. The exact fitted gaps describe the search certificate. For the actual finite-arithmetic learner, use feasible approximate fits as specified below; the potentially exponential number of net points is explicit.

It remains to transfer empirical risks uniformly. Subtract the
loss-dependent constant $G_\Phi(0)=-\min_{q\in\Delta_K}\Phi(q)$.
The normalized canonical loss profile
\[
 \mathcal L^0_{\Phi,y}(q)=\langle q-e_y,\nabla_H\Phi(q)\rangle
                    -\Phi(q)-G_\Phi(0)
\]
has magnitude at most $2\sqrt2M$ and Lipschitz constant
$\sqrt2\lambda$. Indeed, along a segment $q(t)$, the Lipschitz gradient
$a(t)=\nabla_H\Phi(q(t))$ is absolutely continuous, and cancellation
of the product and chain-rule terms gives
\[
 \frac{d}{dt}\mathcal L^0_{\Phi,y}(q(t))
       =\langle q(t)-e_y,a'(t)\rangle
\]
almost everywhere. The simplex diameter is $\sqrt2$, proving the
Lipschitz bound. The normalized comparator profile
$a\mapsto\ell_\Phi(a,y)-G_\Phi(0)$ is $\sqrt2$-Lipschitz and has
magnitude at most $\sqrt2A$ on the score ball, because its subgradients
are differences of two probability vectors.

For a nonempty entropy family, strong convexity implies
$\beta\lambda\ge1$. Both the learned and comparator losses therefore
belong to the class of Lemma~\ref{stat:uniform} with
\[
 C=\sqrt2 B\lambda,
 \qquad M_0=3(M+A),
 \qquad D_0\le cD.
\]
Use the score net with $\eta=\varepsilon/4$. At every net point,
compute a feasible solution of \eqref{stat:socp} with objective gap at most
\[
 \zeta=\min\{1/8,c\varepsilon^2/(A+\lambda)^2\}.
\]
To obtain this fit by finite search, merge equal scores and assign one
probability vector to each distinct score. With only one distinct score
the empirical mean label vector is optimal. Otherwise let $\delta_{\mathrm{sep}}>0$ be the
minimum distance between distinct scores, set $t=\zeta/8$, and choose
the smallest positive integer $N$ such that $\sqrt K/N\le\min\{t,tB\delta_{\mathrm{sep}}/2\}$. Enumerate the
assignments with coordinates in $N^{-1}\mathbb Z$ lying in $\Delta_K$,
retain those satisfying the pairwise constraints, and minimize the empirical
objective with the original sample multiplicities. Break ties by
lexicographic order.
To see the accuracy, mix each exact fitted probability with $u$ in
proportion $t$. Round each mixed vector to the grid simplex, rounding
down all coordinates and distributing the remaining units to the
largest remainders. Each rounding moves its vector by at most
$\sqrt K/N$. Mixing creates slack at least $tB\delta_{\mathrm{sep}}$ in every distinct-score
constraint, so the rounded assignment is feasible. Each probability
moves by at most $(\sqrt2+1)t$ from its exact fit. Since
$p\mapsto\tfrac12\|p-e_y\|^2$ is $\sqrt2$-Lipschitz on the simplex,
the objective increases by less than $\zeta$.
Strong convexity of the half-squared objective gives empirical
probability error at most $a=\sqrt{2\zeta}$. Consequently the computed
gap differs from its exact value by at most $2Aa$.
Choose the matrix with the smallest computed gap. Since the net contains
an exact-gap-$\eta$ candidate, the selected matrix's exact gap is at
most $\eta+4Aa$. Replacing its exact fit by the feasible computed
probabilities adds at most $\sqrt2\lambda a$ to every empirical risk.
Taking $c\le1/512$ makes the total empirical regret at most
$\varepsilon/2$. The same-cap extension applies to these feasible
computed probabilities. Choose $n$ so that the uniform deviation is at
most $\varepsilon/4$, and transfer both forecast and comparator risks.
This proves \eqref{stat:guarantee} and \eqref{stat:sample-bound} with an
actual finite computation. The event is uniform over the entire entropy
family, so the loss may be chosen after training.

\paragraph{Deterministic choices and measurability.}
Project each training feature onto the closed radius-$L$ ball.
This leaves the sample unchanged almost surely. Construct an
orthonormal basis of its span by Gram--Schmidt in sample order,
omitting a vector exactly when its residual norm is zero. For each
possible span dimension use a fixed ordered grid net of the matrix
ball, with the zero-dimensional net consisting of the zero operator.
Run the deterministic feasible fitting routine at the stated accuracy
and choose the first candidate with the smallest computed gap.
These operations and their branch sets are Borel measurable.
The extension in \eqref{stat:extension-qp} is also a measurable
prediction rule: its feasible simplex is fixed and compact, its
objective is continuous in the stored data and query, and all
minimizers give the same probability $P\theta$. Compactness therefore
makes this probability a continuous function of those parameters.
Thus $(S,x)\mapsto p_S(x)$ is jointly Borel measurable. For a separable
Hilbert feature space, the construction uses finitely many sample
inner products and finite linear combinations of sample vectors.

\subsection{Sample lower bounds}
\label{stat:lower-section}
Learners in this subsection have prediction rules jointly measurable in
the training sample, query, and any independent auxiliary random seed.
The lower bounds allow arbitrary probability forecasts and do not
assume a score, norm, or Lipschitz restriction on their outputs.
\begin{proof}[Proof of Proposition~\ref{stat:lower}]
Take $L=R=\beta=\lambda=M=1$, put
$v=(e_1-e_2)/\sqrt2\in H$, and let $X$ be uniform on
$\mathbb S^{d-1}\subset\mathbb R^d$. In particular,
$\mathbb E[XX^\top]=I_d/d$ and the feature law is invariant under
every orthogonal transformation. For a sign vector
$s\in\{-1,1\}^d$, put
\[
 a=\frac1{2K},\qquad w_s=\frac a{\sqrt d}s,
 \qquad \eta_s(x)=u+v\langle w_s,x\rangle.
\]
Every coordinate of $\eta_s(x)$ is at least $1/(2K)$, and the score
matrix $V_s=vw_s^\top$ has Frobenius norm $a\le1$. It is a conditional
Bayes action for the quadratic entropy. Consequently the regret of any
probability forecast $p$ is
\[
 \operatorname{Regret}_s(p)=\frac12\mathbb E_X
                   \|p(X)-\eta_s(X)\|^2.
\]

For a realized output predictor, define
\[
 \widehat w=d\,\mathbb E_X\big[X\langle p(X)-u,v\rangle\big].
\]
The functions $\sqrt d\,X_i$ are orthonormal in $L^2(P_X)$, so
Bessel's inequality gives
\begin{align*}
 \operatorname{Regret}_s(p)
 &\ge\frac12\mathbb E_X
   \big(\langle p(X)-u,v\rangle-\langle w_s,X\rangle\big)^2\\
 &\ge\frac1{2d}\|\widehat w-w_s\|^2\\
 &\ge\frac{a^2}{2d^2}
       \sum_{i=1}^d\mathbf1_{\{\operatorname{sign}(\widehat w_i)\ne s_i\}},
\end{align*}
with any fixed convention for ties. The vector $\widehat w$ is used only to
convert an arbitrary predictor into a sign estimator in the lower bound.

If $s,s'$ differ only in coordinate $i$, then
$\eta_s(X)-\eta_{s'}(X)=2as_iX_i v/\sqrt d$. The elementary bound
$\mathrm{KL}(p\|q)\le\sum_j(p_j-q_j)^2/q_j$ and the coordinate
floor $1/(2K)$ imply that one observation has KL divergence at most
\[
 2K\,\mathbb E_X\|\eta_s(X)-\eta_{s'}(X)\|^2
 =\frac{8Ka^2}{d^2}=\frac2{Kd^2}.
\]
If $n\le Kd^2/16$, Pinsker's inequality therefore bounds the training
distributions' total variation by $1/4$. Pairing sign vectors under a
uniform sign prior gives expected Hamming error at least $3d/8$.
If $N$ is this Hamming error, $0\le N\le d$ yields
\[
 \Pr(N\ge d/8)\ge2/7.
\]
On this event the regret is at least
$a^2/(16d)=1/(64K^2d)$. For sufficiently small $\varepsilon$, choose
$d=\lfloor1/(128K^2\varepsilon)\rfloor$. The event has regret at
least $2\varepsilon$ and the sample threshold is
$\Omega_K(\varepsilon^{-2})$. Averaging over signs provides one fixed
instance with failure probability at least $2/7$. The same argument
includes randomized learners, since their internal randomness cannot
increase total variation.
\end{proof}

\paragraph{Scalar binary normalization.}
The same lower bound applies directly to the monotone binary theorem,
with $L=R=\beta=\lambda=M=1$. Take
$\varphi(q)=(q-1/2)^2/2$ and
$\eta_s(x)=1/2+\langle w_s,x\rangle$, where
$w_s=s/(4\sqrt d)$. Its Bayes action is
$\varphi'(\eta_s(x))=\langle w_s,x\rangle$, so no comparator intercept
is required. The conditional probabilities lie in $[1/4,3/4]$.
One-bit KL is at most $4/(3d^2)$, and the preceding Bessel--Hamming
argument gives regret at least $1/(256d)$ with probability at least
$2/7$ whenever $n\le d^2/16$. Taking
$d=\lfloor1/(512\varepsilon)\rfloor$ proves
$\Omega(\varepsilon^{-2})$ samples at fixed confidence under exact
spherical symmetry, even without restricting the output predictor.

\subsection{Preserving monotonicity in the binary model}
\label{stat:binary-section}
The scalar binary model admits the stronger learned-link constraint of
monotonicity. Here $Y\in\{0,1\}$, scores and actions are real, and
\[
 g_\varphi(a)=\max_{q\in[0,1]}\{qa-\varphi(q)\},
 \qquad \ell_\varphi(a,y)=g_\varphi(a)-ya.
\]
Assume $\varphi$ is continuously differentiable on a neighborhood of $[0,1]$,
is $1/\beta$-strongly convex there on $[0,1]$, and satisfies
$|\varphi'|\le M$ and $\operatorname{Lip}(\varphi')\le\lambda$.
The finite canonical loss is
$\mathcal L_\varphi(q,y)=\ell_\varphi(\varphi'(q),y)$.

\begin{corollary}[A monotone single-index learner]
\label{stat:binary}
For the preceding binary model, let $\|X\|\le L$, $\|w\|\le R$, and
$B\ge\beta$. Define $D_{\mathrm{bin}}=M+LR+B\lambda LR$.
For $0<\varepsilon\le D_{\mathrm{bin}}$, $0<\delta<1/2$, and integer
$n\ge1$ satisfying the bound below, there is a deterministic measurable learner returning
$p(x)=\rho(\langle w,x\rangle)$, with $\|w\|\le R$ and a
nondecreasing $B$-Lipschitz link $\rho:\mathbb R\to[0,1]$, that
simultaneously satisfies
\[
 \mathbb E\mathcal L_\varphi(p(X),Y)
 \le\inf_{\|v\|\le R}\mathbb E\ell_\varphi(\langle v,X\rangle,Y)
       +\varepsilon
\]
for all such $\varphi$, with probability at least $1-\delta$, using
\[
 n\ge C(D_{\mathrm{bin}}/\varepsilon)^2
       \log^2\!\big(CD_{\mathrm{bin}}/(\varepsilon\delta)\big)
\]
samples. No polynomial-time claim is made.
\end{corollary}
\begin{proof}
Fit $Y$ in squared loss by nondecreasing $B$-Lipschitz functions of the
scalar score. This is a compact convex profile class. Every perturbation
needed for Lemma~\ref{stat:empirical} preserves this class: scalar convex
proximal maps are nondecreasing and nonexpansive, clipping is
nondecreasing and nonexpansive, and
$s\mapsto\operatorname{clip}_{[0,1]}((1-t)\rho(s)+tBs)$
is nondecreasing and $B$-Lipschitz. The endpoint residual property
follows by clipping fitted probabilities to $[t,1-t]$ and differentiating
the least-squares inequality, exactly as in the simplex-face argument.
Thus the same certificate, continuity, fixed point, and finite score-net
search apply. The empirical fit is specified by its values at the
ordered sample scores. The finite grid fit above also applies: mix
with $1/2$ and round each scalar value down on a common grid. This
preserves the ordering, and a fine enough grid preserves the slope
constraints by the same slack estimate. Linear interpolation between distinct scores,
with constant continuation outside their range, preserves monotonicity
and the cap $B$ without any enlargement. Finally, the normalized
canonical loss profiles have magnitude at most $2M$ and Lipschitz
constant $\lambda$, and comparator profiles are $1$-Lipschitz.
Apply Lemma~\ref{stat:uniform} in one score dimension, with two label
profiles, to transfer the empirical comparison.
\end{proof}

\subsection{One learned forecast for smooth entropies and logarithmic losses}
\label{stat:log-smoothing-section}

The bounded-gradient assumption excludes negative Shannon entropy from
the rate in Theorem~\ref{stat:pac}. Smoothing the same empirical fit gives
a joint guarantee for that smooth family and a bounded range of
logarithmic losses, at a slower statistical rate. It also produces finite
logits on every feature vector.

For $T>0$, let
\[
 \Phi_T(q)=T\sum_{j=1}^K q_j\log q_j,
 \qquad
 \ell_T(a,y)=T\log\sum_{j=1}^K e^{a_j/T}-a_y.
\]
Fix $0<T_{\min}\le T_{\max}<\infty$ with
$T_{\min}\ge1/(2\beta)$. Then every $\Phi_T$, for
$T\in[T_{\min},T_{\max}]$, belongs to the population family
$\mathcal E_\beta$.

\begin{corollary}[Joint smooth-entropy and log-loss learning]
\label{stat:joint-log-pac}
Assume $\|X\|\le L$, use the radius-$R$ Frobenius comparator ball,
and put $A=LR$. Fix a nonempty smooth entropy family
$\mathcal E(\beta,\lambda,M)$ as in Theorem~\ref{stat:pac}, and a
temperature interval as above. Assume $0<\delta<1/2$ and $0<\varepsilon\le M+A+\beta\lambda A$. Set
\[
 \tau=\min\left\{\frac12,
       \frac{\varepsilon}{8(\sqrt2\lambda+2T_{\max})}\right\},
\]
and define
\[
 D_\tau=M+A+T_{\max}\log(K/\tau)
             +\beta A\big(\lambda+KT_{\max}/\tau\big).
\]
There is a learner returning a single forecast
$p_\tau(x)=\rho_\tau(Wx)$, with
\[
 \|W\|_F\le R,\qquad
 \operatorname{Lip}(\rho_\tau)\le(1-\tau)\beta\le\beta,
 \qquad (p_\tau(x))_j\ge\tau/K,
\]
such that, with probability at least $1-\delta$, it simultaneously
satisfies
\begin{align}
 \mathbb E\mathcal L_\Phi(p_\tau(X),Y)
 &\le\inf_{\|V\|_F\le R}\mathbb E\ell_\Phi(VX,Y)+\varepsilon,
 &&\Phi\in\mathcal E(\beta,\lambda,M),\label{stat:joint-smooth}\\
 \mathbb E[-T\log(p_\tau(X))_Y]
 &\le\inf_{\|V\|_F\le R}\mathbb E\ell_T(VX,Y)+\varepsilon,
 &&T\in[T_{\min},T_{\max}].\label{stat:joint-log}
\end{align}
It suffices to use the sample bound of Theorem~\ref{stat:pac} with
$D$ replaced by $D_\tau$ and a sufficiently large constant $C_K$.
For fixed parameters, this is
$\widetilde O_K(\varepsilon^{-4})$ for $K=2,3$ and
$\widetilde O_K(\varepsilon^{-2(K-1)})$ for $K\ge4$.
\end{corollary}
\begin{proof}
Use exactly the empirical least-squares fits and score-matrix search
from Theorem~\ref{stat:pac}, with cap $B=\beta$. The empirical version
of Lemma~\ref{core:certificate} gives simultaneous empirical regret at
most $\Gamma_n(W)$ for every population entropy in $\mathcal E_\beta$,
including the smooth family and all the logarithmic targets above.
For a fitted probability $p_i$, its realized label is in its support.
Consequently its canonical logarithmic loss is finite.

Apply Lemma~\ref{stat:same-cap-extension} to obtain a head $\rho$ with
cap $\beta$ interpolating the fitted probabilities. Define
\[
 \rho_\tau(s)=(1-\tau)\rho(s)+\tau u.
\]
This has the stated cap and positive coordinate floor. On every
training example, for $T\le T_{\max}$,
\begin{align*}
 -T\log((1-\tau)p_{i,y_i}+\tau/K)
 &\le-T\log p_{i,y_i}-T\log(1-\tau)\\
 &\le-T\log p_{i,y_i}+2T_{\max}\tau.
\end{align*}
For a smooth target, the $\sqrt2\lambda$ Lipschitz bound on its finite
canonical loss and $\|p_i-u\|\le1$ give a smoothing increase of at most
$\sqrt2\lambda\tau$. Thus smoothing costs at most
$\varepsilon/8$ in either family, by the choice of $\tau$.

It remains to transfer these empirical comparisons uniformly. Subtract
$G_\Phi(0)$ from each smooth loss, and subtract $T\log K$ from each
logarithmic loss. The normalized logarithmic profiles have magnitude
at most $T_{\max}\log(K/\tau)$ and, as functions of the unsmoothed
probability, Lipschitz constant at most $KT_{\max}/\tau$. After
composition with $\rho$, their score-profile cap is at most
$\beta KT_{\max}/\tau$. The smooth forecast profiles have cap at most
$\sqrt2\beta\lambda$ and magnitude at most $2\sqrt2 M$.
All normalized comparator profiles have cap at most $\sqrt2$ and
magnitude at most $\sqrt2 A$. Since $\beta\lambda\ge1$, both target
families and their comparators lie in the scalar-profile class of
Lemma~\ref{stat:uniform}, with
\[
 C\le c\beta(\lambda+KT_{\max}/\tau),\qquad
 M_0\le c\big(M+A+T_{\max}\log(K/\tau)\big).
\]
Hence its scale $D_0=M_0+CA$ is at most $cD_\tau$.
Take the score-net tolerance to give $\Gamma_n(W)\le\varepsilon/4$
and the sample size to make both uniform deviations at most
$\varepsilon/8$. Combining the empirical certificate, smoothing bias,
and deviations proves both claims with room for numerical errors.
The event is uniform over both families, and both use the same forecast.

For approximate feasible least-squares fits, compare risks after
smoothing. An objective error $\zeta$ in the half-squared-loss fit
changes the fitted sample probabilities in empirical $L^2$ norm by
at most $\sqrt{2\zeta}$. The gap changes by at most
$2A\sqrt{2\zeta}$, and the smoothed empirical losses change by at most
\[
 (\sqrt2\lambda+KT_{\max}/\tau)\sqrt{2\zeta}.
\]
Thus choosing
\[
 \zeta=\min\left\{\frac18,\ c\,\frac{\varepsilon^2}
                  {(A+\lambda+KT_{\max}/\tau)^2}\right\}
\]
with sufficiently small numerical $c$ preserves the guarantees.
Finite evaluation of the extension is controlled by
\eqref{stat:extension-accuracy}.
Finally, for fixed parameters, $\tau=\Theta(\varepsilon)$ as
$\varepsilon\downarrow0$, so $D_\tau=O(\varepsilon^{-1}
+\log(1/\varepsilon))$, giving the displayed accuracy exponents.
\end{proof}

\paragraph{Finite decision rules and scope.}
For a logarithmic target, the finite Bayes action is
$k_T(p)=T\Pi_H(\log p_1,\ldots,\log p_K)$. The coordinate floor gives
\[
 \|k_T(p_\tau(x))\|
 \le\frac{T_{\max}\sqrt K}{2}\log(K/\tau)
\]
uniformly in $x$. Smoothing does not change the learned score matrix,
and the same probability vector supports every loss in both families.
The finite upper temperature is a genuine normalization assumption for
an additive $\varepsilon$ guarantee. The additional smoothing changes the sample rate because the slope of
the smoothed logarithmic loss grows as the coordinate floor decreases.

\paragraph{A sample lower bound for one logistic loss.}
The $\Omega(\varepsilon^{-2})$ lower bound also holds for the single
binary logistic loss
$\ell(a,y)=\log(1+e^a)-ya$, even when features are uniform on a Euclidean unit sphere
and the learner may output any probability forecast.
To see this, let $X$ be uniform on $\mathbb S^{d-1}$, set
$a=1/4$, and, for $s\in\{-1,1\}^d$, define
\[
 w_s=\frac a{\sqrt d}s,\qquad
 \eta_s(x)=\sigma(\langle w_s,x\rangle),\qquad
 \sigma(z)=\frac1{1+e^{-z}}.
\]
The norm-$a$ linear action $\langle w_s,x\rangle$ is Bayes optimal.
Thus, with $L=R=1$, the comparator benchmark is the Bayes risk, and
the regret of any probability forecast satisfies
\[
 \operatorname{Regret}_s(p)
 =\mathbb E_X\operatorname{KL}
       (\operatorname{Ber}(\eta_s(X))\|\operatorname{Ber}(p(X)))
 \ge2\mathbb E_X(p(X)-\eta_s(X))^2.
\]
Boundary reports are included with the usual infinite-log-loss
convention. Rotational symmetry gives
\[
 d\,\mathbb E_X[X(\eta_s(X)-1/2)]=\kappa_d w_s,
 \qquad \kappa_d\ge1/5.
\]
Indeed, the displayed vector is parallel to $w_s$, its scalar product
with $w_s$ equals $d\mathbb E[S(\sigma(S)-1/2)]$ for
$S=\langle w_s,X\rangle$, and
$S(\sigma(S)-1/2)\ge S^2/5$ because $|S|\le1/4$ and
$\sigma'(z)\ge1/5$ on that interval. Also
$\mathbb E S^2=a^2/d$. The coefficient $\kappa_d$ is the same
for every sign vector because $\|w_s\|=a$.

For any realized predictor put
$\widehat w=d\mathbb E_X[X(p(X)-1/2)]$. Bessel's inequality for the
orthonormal functions $\sqrt d\,X_j$, followed by sign decoding,
gives
\begin{align*}
 \operatorname{Regret}_s(p)
 &\ge\frac2d\|\widehat w-\kappa_dw_s\|^2\\
 &\ge\frac{2\kappa_d^2a^2}{d^2}
       \sum_j\mathbf1_{\{\operatorname{sign}(\widehat w_j)\ne s_j\}}
 \ge\frac{N_s}{200d^2},
\end{align*}
where $N_s$ denotes the Hamming error. If $s,s'$ differ only in
coordinate $j$, the one-observation KL divergence is at most
\[
 \frac18\mathbb E_X
       [\langle w_s-w_{s'},X\rangle^2]
 =\frac{a^2}{2d^2}=\frac1{32d^2}.
\]
Here we used that the Bernoulli logistic log-partition
$g(z)=\log(1+e^z)$ has $g''\le1/4$, and hence its Bregman
divergence is at most one eighth of squared logit distance.
If $n\le4d^2$, Pinsker's inequality gives total variation at most
$1/4$ between the corresponding training laws.
The same sign-pairing argument as in Proposition~\ref{stat:lower}
then gives $\mathbb E N_s\ge3d/8$ under a uniform sign prior,
and therefore $\Pr(N_s\ge d/8)\ge2/7$.
On this event the regret is at least $1/(1600d)$.
Choosing $d=\lfloor1/(3200\varepsilon)\rfloor$ for sufficiently
small $\varepsilon$ proves that success probability greater than
$5/7$ requires $\Omega(\varepsilon^{-2})$ samples.
The conclusion includes randomized learners by the same
total-variation argument. This target is admissible with the fixed
parameters $\beta=1$ and $T_{\min}=T_{\max}=1$.

\section{Proof of the Dirichlet-Geometry Learning Theorem}
\label{app:dirichlet}
Write $\mathcal C=\Delta_K-u$, $Z=e_Y-u$, and
$\mathsf L=G^\dagger$. All arguments below use the entire fixed
feature dictionary, so observed samples need not contain every point.

\subsection{Affine interpolation and contraction}
The rank assumption makes the dictionary affinely independent.
Since $\mathsf L\mathbf1=0$ and its off-diagonal entries are
nonpositive, put $c_{ij}=-\mathsf L_{ij}\ge0$ for $i\ne j$.
For values $t_i$ in any finite-dimensional Euclidean space, form the
matrix $T$ with rows $t_i^\top$. Then
\begin{equation}
 \min_{t_i=b+Ax_i}\|A\|_F^2
 =\operatorname{tr}(T^\top\mathsf L T)
 =\sum_{i<j}c_{ij}\|t_i-t_j\|^2.
 \label{eq:dirichlet-energy}
\end{equation}
Indeed, eliminating $b$ requires
$J_N\mathsf X A^\top=J_NT$. The minimum-norm solution has squared
norm $\operatorname{tr}(T^\top G^\dagger T)$; the rank assumption
ensures solvability, and zero row sums give the edge formula.
For $N=1$ the minimum is zero, attained with a constant interpolant.
Consequently any $C$-Lipschitz transformation of the values increases
this energy by at most a factor $C^2$. In particular, values of
$b+Ax_i$ with $\|A\|_F\le Q$, followed by a $C$-Lipschitz map,
again admit an affine representation with slope norm at most $CQ$.
This applies to vector outputs with a Frobenius norm bound.

\subsection{The exact empirical certificate}
Let $F_n(a,W)$ denote the objective in
\eqref{eq:dirichlet-program}. It has a minimizer because optimizing the
intercept for each $W$ gives
$a=\mathbb E_n Z-\beta W\mathbb E_n X$, a continuous function
on the compact weight ball. Every such intercept has norm at most
$A_*=1+\beta RL$.

After this substitution the objective is continuous and convex on the compact
weight ball. Its minimizers form a compact convex set with a unique
minimum-norm element. This choice depends measurably on the sample. To see this,
add $j^{-1}\|W\|_F^2$ to the reduced objective for $j=1,2,\ldots$.
Each regularized minimizer is unique and continuous in the data by
compactness. Optimality bounds its norm by that of the minimum-norm
minimizer and makes its objective converge to the minimum. Thus every
limit point is the minimum-norm minimizer, which proves convergence
and measurability.
For an optimizer $(a_*,W_*)$, write its raw values on all dictionary
points as $T_i=a_*+\beta W_*x_i$. Their energy is at most
$\beta^2R^2$. Projecting every $T_i$ onto $\mathcal C$ preserves
this bound and, by \eqref{eq:dirichlet-energy}, gives another
feasible raw affine model. For any $z\in\mathcal C$, projection
satisfies
\[
 \|T_i-z\|^2\ge
 \|\Pi_{\mathcal C}(T_i)-z\|^2+
 \|T_i-\Pi_{\mathcal C}(T_i)\|^2.
\]
Thus every observed raw fitted value lies in $\mathcal C$, since
otherwise projection would strictly reduce $F_n$. Unobserved raw
values need not lie there; the returned forecast projects them.

Let the centered fit be $\widetilde P_*=a_*+\beta W_*X$ on the sample, and put
$\xi=\widetilde P_*-Z$ and $b=\E_n\langle \xi,\widetilde P_*\rangle$.
Any nonexpansive transformation of the raw values on the entire
dictionary contracts the energy in~\eqref{eq:dirichlet-energy},
so its transformed values are again realizable by a feasible affine
model. The trimmed-simplex and convex-proximal perturbations from
Lemmas~\ref{core:face}--\ref{core:face-test} therefore apply directly
to this least-squares optimum. In particular, the residual is tangent
to its active simplex face, and for every target entropy
\[
 \E_n L_\Phi(u+\widetilde P_*,Y)
       \le-\E_n\Phi(u+\widetilde P_*)+b/\beta.
\]
First-order optimality in the unrestricted intercept gives $\E_n \xi=0$.
Consequently, with $g_* =\E_n[\xi X^\top]$,
$b=\beta\langle g_*,W_*\rangle_F$.
First-order optimality in the matrix ball gives
\[
 \langle g_*,V-W_*\rangle_F\ge0
       \quad(\|V\|_F\le R),
 \qquad
 b/\beta\le\E_n\langle \xi,VX\rangle.
\]
Fenchel's inequality now bounds every comparator risk below by
$-\E_n\Phi(u+\widetilde P_*)+b/\beta$, proving exact empirical omniprediction.
The support property makes the finite and facewise canonical losses
agree on the observed labels. This proof uses no fixed point.

\subsection{Uniform statistical transfer}
For each target subtract the same constant $G_\Phi(0)$ from its
forecast and comparator losses. With $z_y=e_y-u$, the normalized
canonical loss decomposes as
\begin{align*}
 \mathcal L^0_{\Phi,y}(q)&=h_\Phi(q)-\langle z_y,A_\Phi(q)\rangle,\\
 A_\Phi(q)&=\nabla_H\Phi(q),\\
 h_\Phi(q)&=\langle q-u,A_\Phi(q)\rangle-\Phi(q)-G_\Phi(0).
\end{align*}
The vector map $A_\Phi$ is $\lambda$-Lipschitz and bounded by $M$.
The scalar map $h_\Phi$ is $\lambda$-Lipschitz and bounded in
absolute value by $(1+\sqrt2)M$. For the Lipschitz claim, along
any segment the Lipschitz gradient is absolutely continuous and the
product and chain rules give
$\frac{d}{dt}h_\Phi(q(t))=\langle q(t)-u,
\frac{d}{dt}A_\Phi(q(t))\rangle$ almost everywhere. Use
$\|q-u\|\le1$. The bound on its magnitude follows from
$G_\Phi(0)=-\min\Phi$ and the simplex diameter $\sqrt2$.
Similarly $\mathcal L^0_{\Phi,y}$ is $\sqrt2\lambda$-Lipschitz and bounded
by $2\sqrt2M$: its segment derivative is
$\langle q(t)-e_y,\frac{d}{dt}A_\Phi(q(t))\rangle$.

Consider all forecasts
$p(x)=u+\Pi_{\mathcal C}(a+\beta Wx)$ with $\|W\|_F\le R$.
Apply \eqref{eq:dirichlet-energy}, first to projection and then to
$A_\Phi$ and $h_\Phi$, on the entire dictionary. With
$Q=\lambda\beta R$, there are affine representations
\[
 A_\Phi(p(x_i))=c+Bx_i,\qquad
 h_\Phi(p(x_i))=b+\langle v,x_i\rangle,
\]
such that
\[
 \|B\|_F,\|v\|\le Q,\qquad
 \|c\|\le M+QL,\qquad
 |b|\le(1+\sqrt2)M+QL.
\]
The intercept bounds follow by evaluating at any dictionary point.
Thus every learned loss is contained, on the feature support, in
the class
$b+\langle v,x\rangle-\langle z_y,c+Bx\rangle$
with these coefficient bounds. On any sample, let $\sigma_i$ be
independent uniform signs. The expected norms of the four sums
\[
 \sum_i\sigma_i,\quad
 \sum_i\sigma_i X_i,\quad
 \sum_i\sigma_i z_{Y_i},\quad
 \sum_i\sigma_i z_{Y_i}X_i^\top
\]
are at most $\sqrt n,L\sqrt n,\sqrt n,L\sqrt n$, respectively,
by the second-moment identity and $\|z_y\|\le1$.
Duality of Euclidean and Frobenius norms therefore bounds the
empirical Rademacher complexity by
\[
 \frac{(2+\sqrt2)M+4\lambda\beta RL}{\sqrt n}.
\]
For a comparator, $G_\Phi(a)-G_\Phi(0)$ is $1$-Lipschitz.
Its composition with $Vx$ has an affine representation with slope
norm at most $R$ and intercept magnitude at most $2LR$, by
\eqref{eq:dirichlet-energy}. The remaining term
$-\langle z_y,Vx\rangle$ already has Frobenius radius $R$.
Its normalized loss is bounded by $\sqrt2LR$, and the same
calculation bounds comparator complexity by $cLR/\sqrt n$.

Symmetrization and bounded differences now give simultaneous uniform
deviation at most
\[
 c\left\{\frac{M+LR+\beta\lambda LR}{\sqrt n}
       +(M+LR)\sqrt{\frac{\log(2/\delta)}n}\right\}
\]
for both classes, with probability at least $1-\delta$.
Their suprema are measurable because the feature--label support is finite
and taking the closure of a bounded set of loss-value vectors does
not change these suprema. Choosing the constant in
\eqref{eq:dirichlet-samples} sufficiently large makes each deviation
at most $\varepsilon/2$. The empirical comparison and these two
deviations prove the claimed population guarantee.

Finally let $\rho(s)=u+\Pi_{\mathcal C}(\widehat a+\beta s)$.
It is $\beta$-Lipschitz, and its centered version is the gradient of
the convex function
\[
 \Psi(s)=\frac1\beta\max_{z\in\mathcal C}
       \{\langle\widehat a+\beta s,z\rangle-\tfrac12\|z\|^2\}.
\]
The maximizer is unique and equals
$\Pi_{\mathcal C}(\widehat a+\beta s)$, so Danskin's theorem
gives $\nabla\Psi=\rho-u$. Identifying $H$ with $\R^{K-1}$
gives the stated score count.

\subsection{Matching lower bound on a one-hot dictionary}

\begin{proof}[Proof of Proposition~\ref{prop:dirichlet-lower}]
Use $X$ uniform on $\{e_1,\ldots,e_d\}\subset\R^d$.
Its centered Gram matrix is $J_d$ and its pseudoinverse is also $J_d$,
so the Dirichlet condition holds. Put $a=1/(2K)$,
$v=(e_1-e_2)/\sqrt2\in H$, and for $s\in\{-1,1\}^d$ set
\[
 \eta_s(e_i)=u+\frac{a s_i}{\sqrt d}v,
 \qquad w_s=\frac a{\sqrt d}s,
 \qquad V_s=vw_s^\top.
\]
All probabilities are at least $1/(2K)$, and $\|V_s\|_F=a\le1$.
Thus the quadratic comparator benchmark is the Bayes risk.
Decode $s_i$ from the sign of $\langle p(e_i)-u,v\rangle$.
Each wrong sign contributes at least $a^2/d$ to squared error, so
if $N_s$ signs are wrong, the regret is at least $a^2N_s/(2d^2)$.
For neighboring sign vectors, labels differ only when $X=e_i$.
The one-observation KL divergence is at most $8Ka^2/d^2$,
exactly as in the proof of Proposition~\ref{stat:lower}. Hence for
$n\le Kd^2/16$, the same sign-pairing argument gives
$\Pr(N_s\ge d/8)\ge2/7$ on some instance.
The resulting regret is at least $1/(64K^2d)$.
Taking $d=\lfloor1/(128K^2\varepsilon)\rfloor$ completes the proof.
\end{proof}


\begin{thebibliography}{23}
\providecommand{\natexlab}[1]{#1}
\providecommand{\url}[1]{\texttt{#1}}
\expandafter\ifx\csname urlstyle\endcsname\relax
  \providecommand{\doi}[1]{doi: #1}\else
  \providecommand{\doi}{doi: \begingroup \urlstyle{rm}\Url}\fi

\bibitem[Azagra et~al.(2021)Azagra, {Le Gruyer}, and Mudarra]{azagra2021}
Daniel Azagra, Erwan {Le Gruyer}, and Carlos Mudarra.
\newblock {Kirszbraun}'s theorem via an explicit formula.
\newblock \emph{Canadian Mathematical Bulletin}, 64\penalty0 (1):\penalty0 142--153, 2021.
\newblock \doi{10.4153/S0008439520000314}.

\bibitem[Balakrishnan et~al.(2026)Balakrishnan, Haghtalab, Hsu, Lee, and Zhao]{balakrishnan2026}
Sivaraman Balakrishnan, Nika Haghtalab, Daniel Hsu, Brian~W. Lee, and Eric Zhao.
\newblock Panprediction: Optimal predictions for any downstream task and loss.
\newblock In \emph{Proceedings of the 29th International Conference on Artificial Intelligence and Statistics}, volume 300 of \emph{Proceedings of Machine Learning Research}, pages 4321--4329. PMLR, 2026.
\newblock URL \url{https://proceedings.mlr.press/v300/balakrishnan26a.html}.

\bibitem[Blondel et~al.(2020)Blondel, Martins, and Niculae]{blondel2020}
Mathieu Blondel, Andr{\'e} F.~T. Martins, and Vlad Niculae.
\newblock Learning with {Fenchel--Young} losses.
\newblock \emph{Journal of Machine Learning Research}, 21\penalty0 (35):\penalty0 1--69, 2020.
\newblock URL \url{https://jmlr.org/papers/v21/19-021.html}.

\bibitem[Br{\"u}mmer and {du Preez}(2013)]{brummer2013}
Niko Br{\"u}mmer and Johan {du Preez}.
\newblock The {PAV} algorithm optimizes binary proper scoring rules, 2013.
\newblock URL \url{https://arxiv.org/abs/1304.2331}.
\newblock arXiv:1304.2331.

\bibitem[Gibbs and Tibshirani(2026)]{gibbs2026}
Isaac Gibbs and Ryan~J. Tibshirani.
\newblock Sample-efficient omniprediction for proper losses.
\newblock In \emph{Proceedings of the 39th Conference on Learning Theory}, volume 336 of \emph{Proceedings of Machine Learning Research}, pages 2679--2719. PMLR, 2026.
\newblock URL \url{https://proceedings.mlr.press/v336/gibbs26a.html}.

\bibitem[Gollakota et~al.(2023)Gollakota, Gopalan, Klivans, and Stavropoulos]{gollakota2023}
Aravind Gollakota, Parikshit Gopalan, Adam~R. Klivans, and Konstantinos Stavropoulos.
\newblock Agnostically learning single-index models using omnipredictors.
\newblock In \emph{Advances in Neural Information Processing Systems}, volume~36, 2023.
\newblock \doi{10.52202/075280-0646}.
\newblock URL \url{https://papers.nips.cc/paper_files/paper/2023/hash/2f46ef5725a8eca24f7f24a17955ad1a-Abstract-Conference.html}.

\bibitem[Gopalan et~al.(2022)Gopalan, Kalai, Reingold, Sharan, and Wieder]{gopalan2022}
Parikshit Gopalan, Adam~Tauman Kalai, Omer Reingold, Vatsal Sharan, and Udi Wieder.
\newblock Omnipredictors.
\newblock In \emph{13th Innovations in Theoretical Computer Science Conference (ITCS 2022)}, volume 215 of \emph{Leibniz International Proceedings in Informatics}, pages 79:1--79:21. Schloss Dagstuhl--Leibniz-Zentrum f{\"u}r Informatik, 2022.
\newblock \doi{10.4230/LIPIcs.ITCS.2022.79}.

\bibitem[Gopalan et~al.(2023)Gopalan, Hu, Kim, Reingold, and Wieder]{gopalan2023}
Parikshit Gopalan, Lunjia Hu, Michael~P. Kim, Omer Reingold, and Udi Wieder.
\newblock Loss minimization through the lens of outcome indistinguishability.
\newblock In \emph{14th Innovations in Theoretical Computer Science Conference (ITCS 2023)}, volume 251 of \emph{Leibniz International Proceedings in Informatics}, pages 60:1--60:20. Schloss Dagstuhl--Leibniz-Zentrum f{\"u}r Informatik, 2023.
\newblock \doi{10.4230/LIPIcs.ITCS.2023.60}.

\bibitem[Gopalan et~al.(2024)Gopalan, Okoroafor, Raghavendra, Shetty, and Singhal]{gopalan2024}
Parikshit Gopalan, Princewill Okoroafor, Prasad Raghavendra, Abhishek Shetty, and Mihir Singhal.
\newblock Omnipredictors for regression and the approximate rank of convex functions.
\newblock In \emph{Proceedings of the 37th Conference on Learning Theory}, volume 247 of \emph{Proceedings of Machine Learning Research}, pages 2027--2070. PMLR, 2024.
\newblock URL \url{https://proceedings.mlr.press/v247/gopalan24b.html}.

\bibitem[Hu et~al.(2025)Hu, Tian, and Yang]{hty2025}
Lunjia Hu, Kevin Tian, and Chutong Yang.
\newblock Omnipredicting single-index models with multi-index models.
\newblock In \emph{Proceedings of the 57th Annual ACM Symposium on Theory of Computing (STOC 2025)}, pages 1762--1773. ACM, 2025.
\newblock \doi{10.1145/3717823.3718223}.
\newblock URL \url{https://arxiv.org/abs/2411.13083v2}.
\newblock Full version: arXiv:2411.13083v2.

\bibitem[Hu et~al.(2026)Hu, Tian, and Yang]{hty2026}
Lunjia Hu, Kevin Tian, and Chutong Yang.
\newblock Simultaneous {Blackwell} approachability and applications to multiclass omniprediction.
\newblock In \emph{Proceedings of the 39th Conference on Learning Theory}, volume 336 of \emph{Proceedings of Machine Learning Research}, pages 3593--3634. PMLR, 2026.
\newblock URL \url{https://proceedings.mlr.press/v336/hu26c.html}.

\bibitem[Jordan et~al.(2022)Jordan, M{\"u}hlemann, and Ziegel]{jordan2022}
Alexander~I. Jordan, Anja M{\"u}hlemann, and Johanna~F. Ziegel.
\newblock Characterizing the optimal solutions to the isotonic regression problem for identifiable functionals.
\newblock \emph{Annals of the Institute of Statistical Mathematics}, 74\penalty0 (3):\penalty0 489--514, 2022.
\newblock \doi{10.1007/s10463-021-00808-0}.

\bibitem[Kakade et~al.(2011)Kakade, Kalai, Kanade, and Shamir]{kakade2011}
Sham~M. Kakade, Adam~Tauman Kalai, Varun Kanade, and Ohad Shamir.
\newblock Efficient learning of generalized linear and single index models with isotonic regression.
\newblock In \emph{Advances in Neural Information Processing Systems}, volume~24, pages 927--935, 2011.
\newblock URL \url{https://papers.nips.cc/paper_files/paper/2011/file/30bb3825e8f631cc6075c0f87bb4978c-Paper.pdf}.

\bibitem[Kalai and Sastry(2009)]{kalai2009}
Adam~Tauman Kalai and Ravi Sastry.
\newblock The {Isotron} algorithm: High-dimensional isotonic regression.
\newblock In \emph{Proceedings of the 22nd Conference on Learning Theory}, 2009.
\newblock URL \url{https://www.learningtheory.org/colt2009/papers/001.pdf}.

\bibitem[Kirszbraun(1934)]{kirszbraun1934}
Moj{\.{z}}esz~David Kirszbraun.
\newblock {\"U}ber die zusammenziehende und {Lipschitzsche} {Transformationen}.
\newblock \emph{Fundamenta Mathematicae}, 22:\penalty0 77--108, 1934.
\newblock \doi{10.4064/fm-22-1-77-108}.

\bibitem[Lu et~al.(2025)Lu, Roth, and Shi]{lu2025}
Jiuyao Lu, Aaron Roth, and Mirah Shi.
\newblock Sample efficient omniprediction and downstream swap regret for non-linear losses.
\newblock In \emph{Proceedings of the 38th Conference on Learning Theory}, volume 291 of \emph{Proceedings of Machine Learning Research}, pages 3829--3878. PMLR, 2025.
\newblock URL \url{https://proceedings.mlr.press/v291/lu25b.html}.

\bibitem[Martins and Astudillo(2016)]{martins2016}
Andr{\'e} Martins and Ramon Astudillo.
\newblock From softmax to sparsemax: A sparse model of attention and multi-label classification.
\newblock In \emph{Proceedings of the 33rd International Conference on Machine Learning}, volume~48 of \emph{Proceedings of Machine Learning Research}, pages 1614--1623. PMLR, 2016.
\newblock URL \url{https://proceedings.mlr.press/v48/martins16.html}.

\bibitem[McDiarmid(1989)]{mcdiarmid1989}
Colin McDiarmid.
\newblock On the method of bounded differences.
\newblock In J.~Siemons, editor, \emph{Surveys in Combinatorics, 1989}, volume 141 of \emph{London Mathematical Society Lecture Note Series}, pages 148--188. Cambridge University Press, 1989.
\newblock \doi{10.1017/CBO9781107359949.008}.

\bibitem[Noarov et~al.(2025)Noarov, Ramalingam, Roth, and Xie]{noarov2025}
Georgy Noarov, Ramya Ramalingam, Aaron Roth, and Stephan Xie.
\newblock High-dimensional prediction for sequential decision making.
\newblock In \emph{Proceedings of the 42nd International Conference on Machine Learning}, volume 267 of \emph{Proceedings of Machine Learning Research}, pages 46762--46783. PMLR, 2025.
\newblock URL \url{https://proceedings.mlr.press/v267/noarov25b.html}.

\bibitem[Noarov and Roth(2026)]{noarov2026}
Georgy Noarov and Aaron Roth.
\newblock Optimal deterministic multicalibration and omniprediction, 2026.
\newblock URL \url{https://arxiv.org/abs/2606.20557}.
\newblock arXiv:2606.20557.

\bibitem[Okoroafor et~al.(2025)Okoroafor, Kleinberg, and Kim]{okoroafor2025}
Princewill Okoroafor, Robert Kleinberg, and Michael~P. Kim.
\newblock Near-optimal algorithms for omniprediction.
\newblock In \emph{Proceedings of the 66th IEEE Annual Symposium on Foundations of Computer Science (FOCS 2025)}, pages 1595--1609. IEEE, 2025.
\newblock URL \url{https://arxiv.org/abs/2501.17205}.
\newblock Full version: arXiv:2501.17205.

\bibitem[Shamir(2015)]{shamir2015}
Ohad Shamir.
\newblock The sample complexity of learning linear predictors with the squared loss.
\newblock \emph{Journal of Machine Learning Research}, 16\penalty0 (108):\penalty0 3475--3486, 2015.
\newblock URL \url{https://www.jmlr.org/papers/v16/shamir15a.html}.

\bibitem[Zarifis et~al.(2024)Zarifis, Wang, Diakonikolas, and Diakonikolas]{zarifis2024}
Nikos Zarifis, Puqian Wang, Ilias Diakonikolas, and Jelena Diakonikolas.
\newblock Robustly learning single-index models via alignment sharpness.
\newblock In \emph{Proceedings of the 41st International Conference on Machine Learning}, volume 235 of \emph{Proceedings of Machine Learning Research}, pages 58197--58243. PMLR, 2024.
\newblock URL \url{https://proceedings.mlr.press/v235/zarifis24a.html}.

\end{thebibliography}
\end{document}